\documentclass{article}

\PassOptionsToPackage{numbers,sort&compress}{natbib}
\usepackage[main,final]{neurips_2026}
\usepackage[utf8]{inputenc}
\usepackage[T1]{fontenc}
\usepackage{hyperref}
\usepackage{url}
\usepackage{booktabs}
\usepackage{amsmath}
\usepackage{amsfonts}
\usepackage{amssymb}
\usepackage{nicefrac}
\usepackage{microtype}
\usepackage{xcolor}
\usepackage{todonotes}
\usepackage{titletoc}
\usepackage{enumitem}
\usepackage{fontawesome5}
\usepackage{twemojis}
\usepackage{tcolorbox}

\definecolor{accent}{HTML}{4B44CE}
\definecolor{accentsoft}{HTML}{EFEEFC}

\newcommand{\projectlinks}{
\begin{center}
\small
\hypersetup{pdfborder={0 0 0}}
\textcolor{accent}{\textbf{Project Page:}}
\href{https://applewpj.github.io/stepwise-token-learning}{
  \texttt{\textcolor{accent}{https://applewpj.github.io/stepwise-token-learning}}
}\\[0.35em]
\href{https://applewpj.github.io/stepwise-token-learning}{\tcbox[on line, colback=accentsoft, colframe=accentsoft, boxrule=0pt, arc=6pt, boxsep=0pt, left=8pt, right=8pt, top=2.5pt, bottom=2.5pt]{\textcolor{accent}{\faGamepad\hspace{0.35em}\textbf{Come play with our interactive figures!}}}}
\end{center}
}

\title{Smooth Scaling Laws Hide Stepwise Token Learning}
\author{
    Pingjie Wang\thanks{Equal contribution.}\,\,\thanks{Work done during an internship at Dots Studio, Xiaohongshu Inc.} \\
    Dots Studio, Xiaohongshu Inc. \\
    Shanghai Jiao Tong University \\
    \texttt{applewpj@gmail.com} \\
    \And
    Zechen Hu\footnotemark[1]\,\,\thanks{Corresponding author.} \\
    Dots Studio, Xiaohongshu Inc. \\
    \texttt{huzechen@outlook.com} \\
    \And
    Peiru Yang\footnotemark[1]\,\,\footnotemark[2] \\
    Dots Studio, Xiaohongshu Inc. \\
    Tsinghua University \\
    \texttt{ypr21@mails.tsinghua.edu.cn} \\
    \And
    Fu Guo \\
    Dots Studio, Xiaohongshu Inc. \\
    \texttt{guofu@xiaohongshu.com} \\
    \And
    Debing Zhang\footnotemark[3] \\
    Dots Studio, Xiaohongshu Inc. \\
    \texttt{debingzhangchina@gmail.com} \\
}

\begin{document}

\maketitle

\vspace{-0.5cm}
\projectlinks

\begin{abstract}
Language model loss follows remarkably regular scaling laws over model and data size, yet it remains unclear why the aggregate loss should exhibit a power-law form.
Existing explanations often attribute this regularity to a heavy-tailed spectrum of pattern difficulty in natural language, but this view has not been directly validated at token-level granularity in large-scale real-data training.
We present a token-level framework that decomposes scaling laws into localized learning events of individual contextualized tokens.
By fitting token loss trajectories with sigmoids, we show that token learning is concentrated in localized transitions, giving rise to a learning-time spectrum that dominates the scaling-law shape.
Across more than one hundred pre-training runs on large and diverse real-language corpora with modern LLM architectures, scaling up to 6B parameters and 300B training tokens, the measured learning-time spectrum quantitatively reconstructs the validation loss derivative along the training-step $T$, data-scale $D$, and model-scale $M$ axes.
We further show that the same signal is actionable: by reshaping the training distribution according to when tokens become learnable, we alter the optimization trajectory and achieve 11\% faster validation-loss reduction.
These results provide direct empirical evidence that scaling laws are governed primarily by the distribution of token-level learning times, and that this distribution can be used not only to explain scaling behavior but also to improve training performance.
\end{abstract}

\section{Introduction}

\begin{figure}[t]
    \centering
    \includegraphics[width=\linewidth]{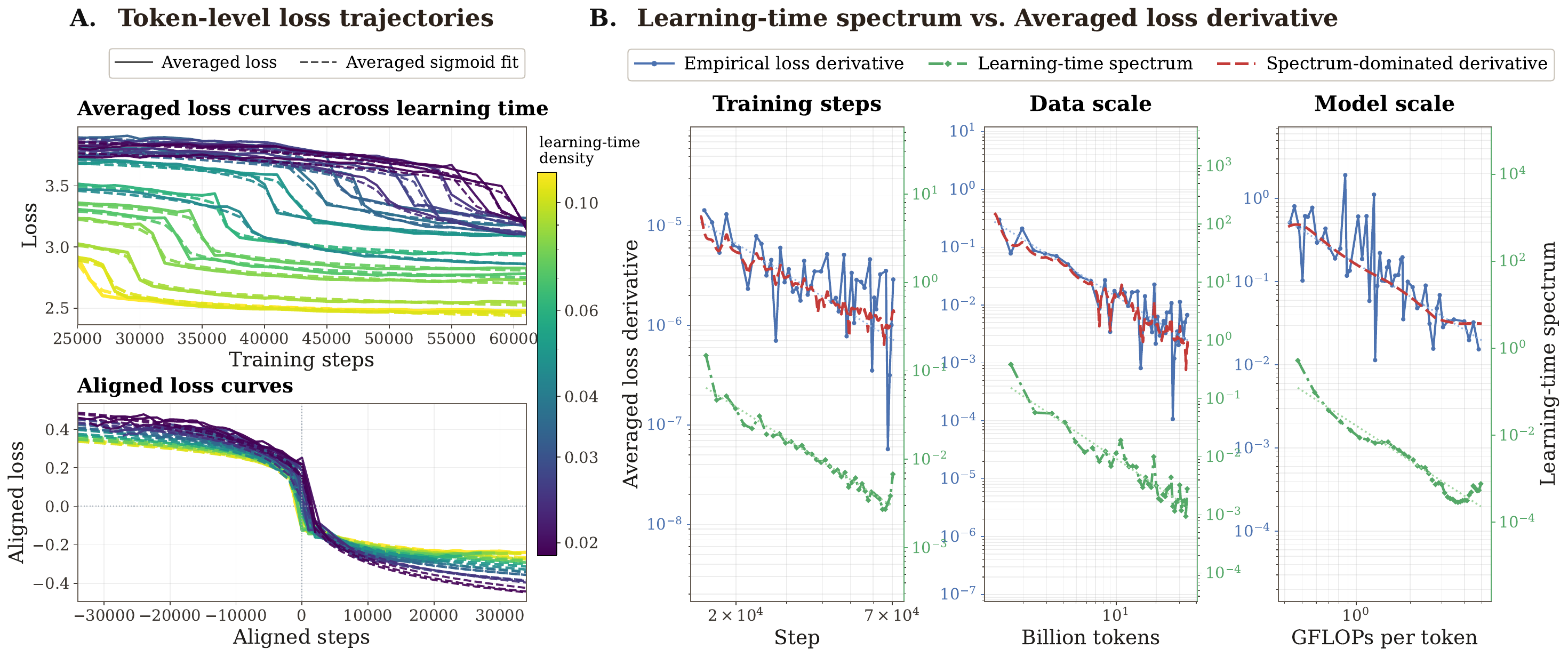}
    \caption{\textbf{A.} Token-level loss trajectories.
    Individual token losses are well fit by sigmoids (\textit{top}), showing that token learning is concentrated in localized transitions at specific learning times rather than spread uniformly over training.
    After aligning tokens by learning time, their loss curves collapse to a highly similar shape (\textit{bottom}).
    \textbf{B.} Learning-time spectrum and empirical loss derivative. 
    The learning-time spectrum is heavy-tailed and approximately power-law, and the loss derivative reconstructed from the measured learning-time spectrum together with the shared learning pulse closely matches the empirical one along the training-step $T$, data-scale $D$, and model-scale $M$ axes.
    These results indicate that the power-law form of the global loss is governed primarily by the distribution of token learning times, rather than by individual token trajectories.}
    \label{fig:first-figure}
\end{figure}

Scaling laws show that language model loss and downstream performance follow predictable trends as model size, data, and training compute increase~\cite{kaplan2020scaling, hoffmann2022training}.
In recent large-scale training practice, they have directly guided compute allocation and motivated major shifts in training strategy, especially toward compute-optimal parameter-data tradeoffs~\cite{hoffmann2022training, gadre2025language}.
However, a basic question remains: \textbf{\emph{Why should the aggregate loss of a language model follow a power law}}?

To explain this phenomenon, recent work attributes the power-law scaling to the spectrum of patterns inherent in the training data~\cite{hutter2021learning,michaud2023quantization,maloney2022solvable,brill2024neural,cagnetta2025learning}.
The intuition is that natural language does not consist of uniformly difficult features; it exhibits a long-tail distribution spanning from high-frequency, low-order linguistic rules to low-frequency, high-order complexities~\cite{piantadosi2014zipf, linders2023zipf}.
From this perspective, the training process acts as a progressive coverage of this spectrum, where models first master dominant patterns before capturing rarer, more sophisticated dependencies~\cite{zhang2021you, rahaman2019spectral, xu2020curriculum}.
Under this view, the power-law scaling is rooted in intrinsic statistical structure of natural language itself.
However, despite the broad appeal of this data-distribution perspective, it has not yet been decomposed at a sufficiently fine-grained level or directly validated in large-scale real-data training runs with industrial-scale models.

In this work, we introduce a token-level perspective to characterize the relationship between token-level loss dynamics, data distribution structure, and aggregate model loss, thereby unraveling the origin of power-law scaling.
Our main contribution is a direct empirical demonstration on real-world data that the power-law form of the global loss fundamentally arises from the statistical distribution of token-level learning events.
As illustrated in Fig.~\ref{fig:first-figure} (A, \textit{top}), individual token losses do not decrease uniformly: they remain on a plateau before and after a localized transition, a pattern that is accurately captured by sigmoid fits.
When tokens are grouped by their learning times (sigmoid center), the resulting learning-time spectrum is long-tailed and follows a power law itself: many tokens are learned early, and progressively fewer tokens learned at later stages.
At the same time, Fig.~\ref{fig:first-figure} (A, \textit{bottom}) shows that after aligning tokens by their learning times, the local shapes of their loss drops are highly similar.
These two observations suggest a simple explanation for the global loss curve: the macroscopic power-law scaling behavior should be governed primarily by the spectrum of learning times, i.e., by how many tokens are learned at each stage.

We validate our theory on large-scale real-world corpora and modern Large Language Model (LLM) architectures used in industrial pre-training pipelines.
Our experiments span models from 290M to 6B parameters and training budgets from 1B to 300B tokens, comprising over 110 runs using 1,178 A100 GPU-days, scaling along three axes: training-step $T$, data-scale $D$, and model-scale $M$.
As summarized in Fig.~\ref{fig:first-figure} (B), taking the derivative makes each token's loss drop visible as a localized learning pulse, indicating when that token is being learned.
The measured learning-time spectrum, combined with the shared pulse, can reconstruct the empirical loss derivative along all $T$, $D$, and $M$ axes.
Prior work has explored improving LLM pre-training by macroscopically adjusting data mixtures or schedules~\cite{anonymous2026small, wang2026mining, albalak2023efficient, zhang2026beyond}, but such interventions are often guided by heuristic signals rather than a fine-grained theory.
Building on the decomposition, we further use the learning-time signal to reshape the training distribution, upweighting samples whose token-level learning events concentrate in a target training interval and downweighting samples that contribute little to that interval.
This intervention changes the subsequent optimization trajectory and yields a 11\% acceleration in validation-loss reduction relative to the original distribution, showing that the learning-time spectrum is not only explanatory but also practical for controlling scaling behavior.
Our main findings are summarized as follows:
\begin{itemize}
\item \textbf{Token learning is non-uniform and transition-like.} Individual token losses remain on a plateau before and after a localized transition, a pattern well captured by sigmoid fits.
\item \textbf{Learning pulse shape is shared across learning times.} Each token loss's derivative forms a localized learning pulse, and they share a similar shape after alignment by learning time.
\item \textbf{Scaling-law is dominated by learning-time spectrum.} The validation loss is driven by the learning-time distribution, i.e., by how many tokens are learned at each axis location.
\item \textbf{The scaling behavior can be reshaped.} Measured learning times can be used to reshape the training distribution, change the subsequent optimization trajectory, and therefore accelerate validation-loss reduction.
\end{itemize}

\section{Related Work}

\paragraph{Empirical Scaling Laws in Language Modeling.}

In language modeling, validation loss scales as a power law with model size, data size, and training compute~\cite{kaplan2020scaling}.
This empirical regularity has become a practical guide for compute allocation and parameter-data tradeoffs in large-scale training~\cite{hoffmann2022training}.
Subsequent work further investigates the optimal allocation of compute resources to maximize model performance, alongside exploring scaling behaviors under specific conditions~\cite{porian2024resolving, chen2025revisiting, gadre2025language}.
Beyond final convergence regarding model size $N$ and data size $D$, the continuous training trajectory itself exhibits a predictable power-law decay with respect to the number of training steps $T$~\cite{kaplan2020scaling, li2025functional, luo2025multi}.

\paragraph{Data-Distribution Explanations of Scaling Laws.}

Existing explanations for scaling laws largely attribute them to the long-tailed structure of the data distribution, suggesting that the observed power laws reflect a long-tail spectrum of learnable patterns or latent features in the data~\cite{hutter2021learning, maloney2022solvable, brill2024neural}.
For example, \citet{hutter2021learning} show that a Zipfian data distribution induces a power-law learning curve, with the exponent set by the Zipf exponent.
Furthermore, several works support this view through toy models and synthetic data experiments~\cite{michaud2023quantization, allen2023physics, cagnetta2025learning, barkeshli2026origin}.
In particular, \citet{michaud2023quantization} use an MLP toy model on a Zipf-distributed sparse parity synthetic dataset to show that neural scaling can be decomposed into the learning of distinct tasks.
By employing different types of synthetic data structures, \citet{cagnetta2025learning} study power-law learning curves using PCFG-generated hierarchical compositional data, whereas \citet{barkeshli2026origin} use synthetic data from function learning on random graphs.
Taken together, these works form a broad line of explanation that links scaling laws to properties of the data distribution, such as compressibility, spectral structure, hierarchy, and distributions of learning complexity.
However, most existing studies remain at the level of theoretical analysis or coarse-grained macroscopic statistics, without directly characterizing scaling behavior at the token level.
Moreover, many of these analyses are conducted in toy settings, with limited validation on large-scale real-world data or modern model architectures used in industrial practice.

\paragraph{Token-Level Training Dynamics.}

Recent work has also highlighted the importance of token-level training dynamics by tracking how individual token instances are learned during training.
Several studies provide qualitative evidence that token-level loss trajectories are highly heterogeneous, with different tokens exhibiting markedly different learning patterns~\cite{carlini2022quantifying, biderman2023pythia, xia2023training, lin2024not, chang2024characterizing, michaelov2025language}.
For example, \citet{lin2024not} track the loss dynamics of a large number of tokens during training and show that token losses do not decrease uniformly, but instead separate into components associated with effective learning and noise.
\citet{chang2024characterizing} further show that the learning trajectories of individual token instances are reproducible across training runs, suggesting that token learning is shaped by sequential learning dependencies. 
They also relate these dynamics to text-level statistical features.
Together, these works show that learning is highly heterogeneous at the token level.
However, they mainly provide qualitative observations of these dynamics, and do not explicitly model how such micro-level heterogeneity aggregates into the global scaling laws observed at the loss level.
In contrast, our work directly connects token-level training dynamics to scaling laws, and validates this connection on modern LLMs with industry-scale data and architectures.
\section{Decomposing Scaling Laws}
\label{sec:method}

\subsection{From Macroscopic Power Laws to Token-Level Decomposition}

Scaling laws are typically expressed at the level of aggregate loss: along axis $a$, validation loss follows a regular power-law decay of the form $L(a)=ka^{-\alpha}+E$~\cite{kaplan2020scaling, hoffmann2022training}, where $a$ can be training steps $T$, data scale $D$, and model scale $M$. 
This leads to a central question: \textbf{\emph{What microscopic mechanism gives rise to the power-law decay of aggregate loss}}?
Three natural hypotheses about what drives the power law are:
\begin{itemize}[leftmargin=*]
\item \emph{Hyp. A}: it comes mainly from overall training dynamics, such as optimization, scheduling, or noise.
\item \emph{Hyp. B}: it comes from individual token-loss trajectories themselves decreasing as power laws.
\item \emph{Hyp. C}: it comes from different tokens being learned at different points along the scaling axis.
\end{itemize}

To distinguish these possibilities, we decompose the validation loss $L(a)$ back to token-level: $L(a) = \frac{1}{|\mathcal I|} \sum_{i \in \mathcal I} \ell_i(a)$.
$\mathcal I$ is the set of contextualized token instances from the validation set, and $\ell_i(a)$ is the loss of token instance $i$ measured along axis $a$.
More directly, the loss derivative tracks where learning occurs: $L'(a) := -\frac{dL(a)}{da} = -\frac{1}{|\mathcal I|} \sum_{i \in \mathcal I} \frac{d\ell_i(a)}{da}$.
Therefore, our core task is to decompose $L'(a)$ into token-level loss derivative and identify which factor truly governs its macroscopic shape.

\subsection{Parameterizing Token-Level Learning Events}

\begin{figure}[t]
    \centering
    \includegraphics[width=\linewidth]{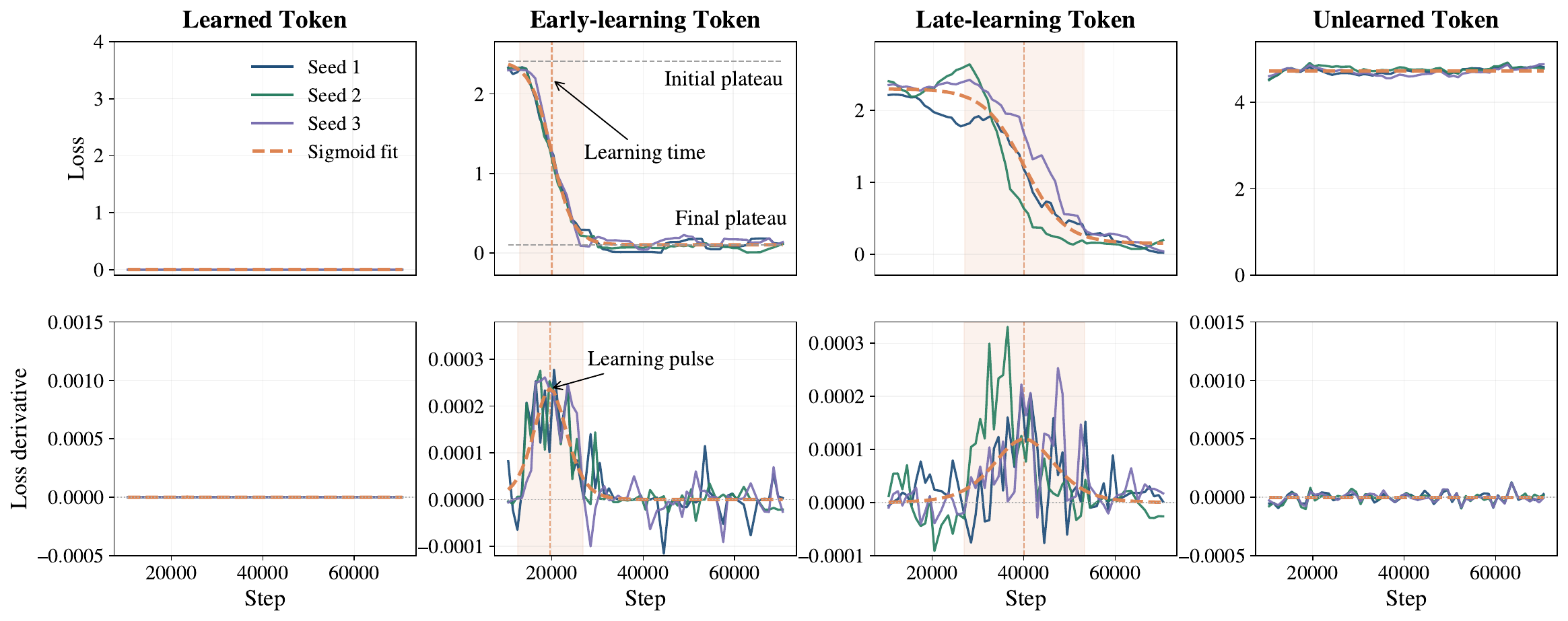}
    \caption{Representative token-level loss trajectories across three independent runs and the corresponding sigmoid fits. The sigmoid fits capture the dominant learning event for each token, which is characterized by a localized sharp descent that occurs at a specific learning time. This suggests that token learning occurs as concentrated events rather than uniformly spread across the axis. Based on the location of learning time, tokens can be grouped into 4 categories: learned, early-learning, late-learning, and unlearned tokens.}
    \label{fig:token-level-case}
\end{figure}

We begin with an observational study comparing the trajectories of the same token instances across independent runs with different random seeds.
Representative cases are shown in Fig.~\ref{fig:token-level-case}.

Three observations consistent with prior works on token-level learning dynamics~\cite{xia2023training,lin2024not, chang2024characterizing} are visible:
Firstly, the same token exhibits highly similar loss descent timing and trajectory shape across different runs, indicating that its learning behavior is determined mainly by token-level structure rather than by optimizer noise, which weakens Hypothesis A.
Secondly, different token losses do not individually follow power-law decay, which argues against Hypothesis B.
Thirdly, the key difference across token instances is when their loss starts to drop, as tokens begin their sharp loss decrease at distinct stages of training.
As shown in Fig.~\ref{fig:token-level-case}, the token-level loss typically begins in a high initial plateau, where the model cannot yet predict the token reliably.
Once the model learns the relevant predictive pattern, the token loss drops rapidly over a localized interval, after which it enters a lower-loss plateau.
Thus, token learning is not uniformly spread across the axis, but occurs as a concentrated \emph{learning event}.
\citet{nam2024exactly} model the emergence of task-level abilities~\cite{wei2022emergent} on synthetic data with sigmoid functions.
Motivated by our observations above, we further model real-world token-level learning events with the following sigmoid parameterization:
\begin{equation}
\ell_i(a)
\approx \ell_i^{\mathrm{post}}
+ \Delta_i \left[1 - \sigma\!\left(\lambda_i(a-\tau_i)\right)\right],
\label{eq:method-sigmoid}
\end{equation}
where $\sigma(z)=1/(1+e^{-z})$, and $\Delta_i := \ell_i^{\mathrm{pre}} - \ell_i^{\mathrm{post}}$.
The parameters in this modeling have direct interpretations: $\tau_i$ is the \emph{learning time}, the center of the dominant descent.
$\lambda_i$ controls the sharpness of the transition and therefore the temporal width of the learning event,
$\Delta_i$ is the total loss reduction contributed by this event,
$\ell_i^{\mathrm{pre}}$ and $\ell_i^{\mathrm{post}}$ are the initial and final plateaus.
Importantly, we use the sigmoid not to claim exact logistic dynamics, but as a simple, interpretable model for extracting the dominant learning event from noisy token-level trajectories.

The loss derivative of token $i$ is:
$\ell_i'(a)
:= -\frac{d\ell_i(a)}{da}
\approx \Delta_i \lambda_i \,
\sigma\!\left(\lambda_i(a-\tau_i)\right)
\left[1-\sigma\!\left(\lambda_i(a-\tau_i)\right)\right]$.
Using $\sigma(z)(1-\sigma(z))=\frac{1}{4}\,\mathrm{sech}^2(z/2)$, this can be equivalently written as
\begin{equation}
\ell_i'(a)
\approx \frac{\Delta_i \lambda_i}{4}\,
\mathrm{sech}^2\!\left(\frac{\lambda_i(a-\tau_i)}{2}\right).
\label{eq:method-pulse-sech}
\end{equation}
We denote $\ell_i'(a)$ as the token's \emph{learning pulse}.
It is centered at $\tau_i$, has width on the order of $\lambda_i^{-1}$, and integrates to the total loss reduction: $\int_{-\infty}^{\infty} \ell_i'(a)\, da = \Delta_i$.
In other words, $\Delta_i$ identifies \emph{how much} is learned, $\lambda_i$ identifies \emph{how abruptly} it is learned, and $\tau_i$ shows \emph{when} it is learned.
Such a localized learning pulse is visualized in Fig.~\ref{fig:token-level-case} (\emph{bottom}).

Under this model, Eq.~\ref{eq:method-pulse-sech} shows that a token contributes to aggregate loss reduction only in a neighborhood of its learning time.
At any axis location $a$, tokens with $\tau_i$ close to $a$ contribute most of the descent rate, while tokens with $\tau_i$ far from $a$ contribute little because they are either not yet undergoing their main learning event or have already completed it.
Therefore,
\begin{equation}
L'(a) = \frac{1}{|\mathcal I|} \sum_{i \in \mathcal I} \ell_i'(a) \approx \frac{1}{|\mathcal I|} \sum_{i \in \mathcal I} \frac{\Delta_i \lambda_i}{4}\,
\mathrm{sech}^2\!\left(\frac{\lambda_i(a-\tau_i)}{2}\right).
\label{eq:method-rate-sum}
\end{equation}
This admits a direct interpretation as the superposition of the token-level learning events that are active around $a$, which is the bridge to the decomposition developed next.

\subsection{Decoupling Learning Shape and Time}

The decomposition in Eq.~\ref{eq:method-rate-sum} suggests that the aggregate loss derivative is formed by many token-level learning pulses. 
Each pulse contains two pieces of information: its local pulse shape (determined by $\Delta_i$ and $\lambda_i$), which describes how the loss decreases once the token is being learned, and its pulse center (determined by $\tau_i$), which specifies when this decrease occurs. 

To explain the source of the macroscopic power-law, we need to separate these two effects.
We define $\tau_i$ as the \emph{learning time} of token $i$, which is the center of the token's \emph{learning pulse} $\ell_i'(a)$, both of which are clearly exhibited in Fig.~\ref{fig:token-level-case}.
Under this view, $L'(a)$ is the superposition of token-local pulses occurring at different learning times $\tau_i$.
To isolate the shape component, we remove the timing variation by shifting each token's pulse so that its center is placed at zero. 
This gives the average centered learning pulse
\begin{equation}
g(a) := \frac{1}{|\mathcal I|}\sum_{i \in \mathcal I} \ell_i'(a+\tau_i),
\label{eq:method-average-pulse}
\end{equation}
which serves as a typical template for the local shape of a token-level learning event.
Once the local pulse shape is summarized by $g$, what remains is to describe how the pulse centers $\tau_i$ are distributed over the axis. 
We capture this timing component with the empirical learning-time spectrum
\begin{equation}
p(\tau) := \frac{1}{|\mathcal I|}\sum_{i \in \mathcal I}\delta(\tau-\tau_i),
\label{eq:method-spectrum}
\end{equation}
where $\delta(\cdot)$ is the Dirac delta. 
Together, $g$ and $p$ separate the two factors that were mixed in Eq.~\ref{eq:method-rate-sum}: $g$ describes the typical local pulse shape, while $p$ describes how these pulses are distributed over training time.
If the aligned pulses have similar shapes that $\ell_i'(a) \approx g(a-\tau_i)$, each token-level pulse can be approximated as the same template shifted to its learning time.
Then Eq.~\ref{eq:method-rate-sum} becomes a superposition of shifted copies of $g$:
\begin{equation}
L'(a)
\approx \frac{1}{|\mathcal I|}\sum_{i \in \mathcal I} g(a-\tau_i)
\approx \int p(\tau)\, g(a-\tau)\, d\tau
= (p * g)(a).
\label{eq:method-convolution}
\end{equation}

Under this decomposition, $g(a)$ captures the local shape of a typical learning event, while $p(\tau)$ captures how many learning events occur at each axis location.
\textbf{\emph{If $g$ is approximately shared across tokens, the macroscopic shape of $L'(a)$ is governed primarily by the \emph{learning-time spectrum} $p(\tau)$ rather than by token-specific loss shape.}}
This is the hypothesis to be tested in our experiments.

Moreover, while $p(\tau)$ is defined formally here, the hierarchical and compositional structure of natural language provides a linguistic reason for it to follow a power-law distribution, because such regularities create a long-tailed spectrum of learning times, as discussed in Appendix~\ref{app:token_difficulty_powerlaw}.
We also compare $p(\tau)$ with existing data-centric curriculum proxies and pre-training intuitions, including data domain, token frequency, and n-gram dependency length. 
The results show that $p(\tau)$ is consistent with these macroscopic data strategy consensuses, suggesting that the learning-time spectrum provides a more fundamental explanation for why such proxies are useful; details are provided in Appendix~\ref{app:align-tau-consensus}.

\section{Step-Axis Decomposition}
\label{sec:step}

\begin{figure}[t]
    \centering
    \includegraphics[width=\linewidth]{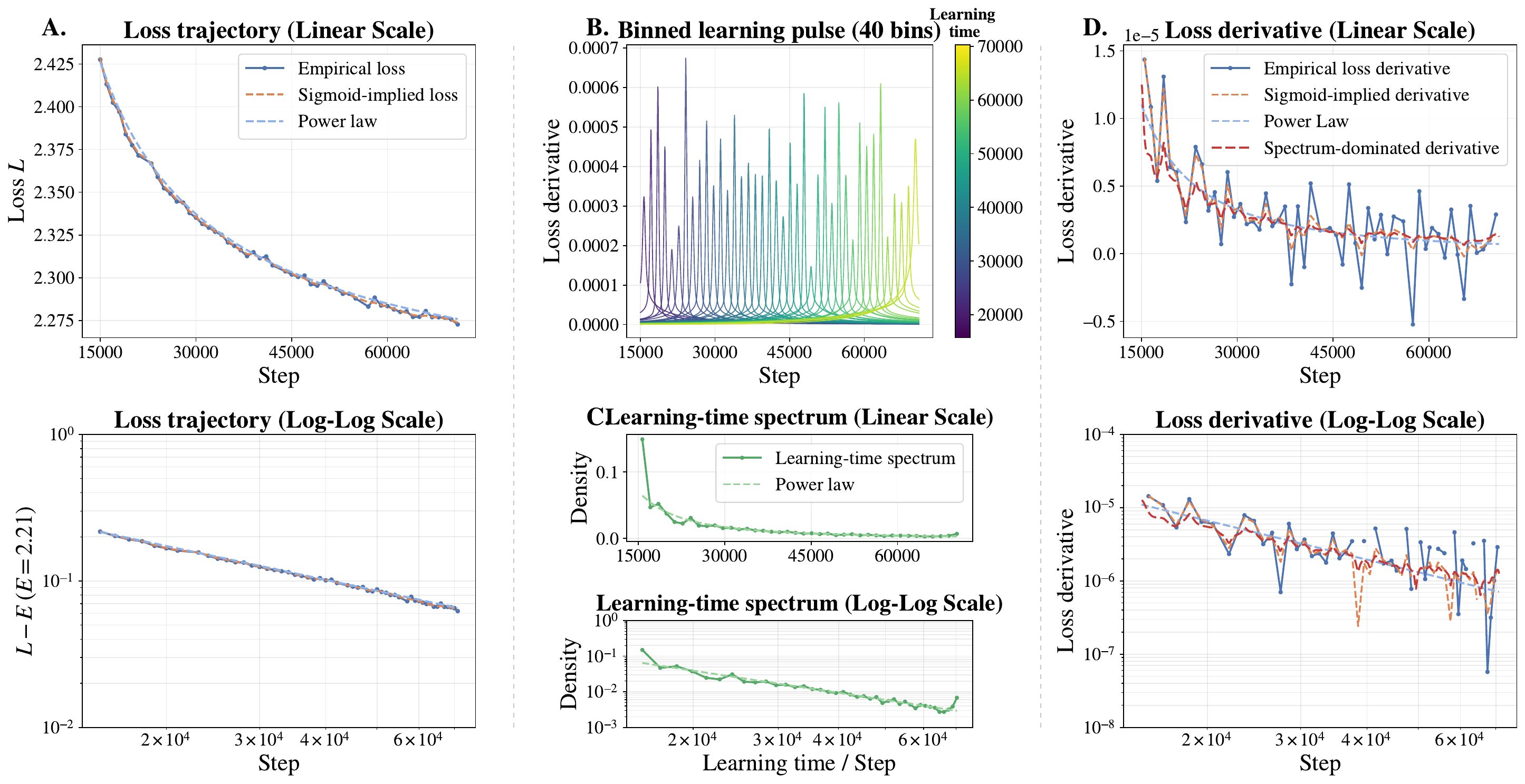}
    \caption{\textbf{A.} Empirical validation loss on the step axis and the aggregate loss implied by token-wise sigmoid fits for a 1.5B model trained on 300B tokens. Both exhibit a clear power-law regime after the initial transient stage. \textbf{B.} Learning pulses estimated from tokens grouped into 40 bins by learning time; after grouping by learning time, the pulse shapes remain highly similar across bins. \textbf{C.} Empirical learning-time spectrum, which is strongly heavy-tailed and is well approximated by a power law. \textbf{D.} Empirical loss derivative, sigmoid-implied derivative, power-law fit, and spectrum-dominated reconstruction on the step axis. Their close agreement shows that the shape of the macroscopic loss derivative is determined primarily by the learning-time spectrum rather than by fine-grained variation in token-level transition shape.}
    \label{fig:step-axis}
\end{figure}

\subsection{Experimental Setup}

We validate the pulse-spectrum mechanism in two stages. We first test it on controllable synthetic data, where the underlying difficulty distribution can be specified explicitly and the full decomposition can be verified cleanly; these results are deferred to Appendix~\ref{app:synthetic-results}. We then turn to industrial-scale real-data pre-training and begin on the training-step axis $T$, since token-level learning events unfold most directly over optimization time and prior work has shown that loss also exhibits regular power-law-like decay over training steps~\cite{kaplan2020scaling,hestness2017deep,luo2025multi,choshen2025hitchhiker,bordelon2024dynamical}. The step axis is therefore the natural first testbed for establishing the full token-to-loss decomposition before extending the analysis to $D$ and $M$.

For the step-axis analysis, we train a 1.5B Dense model on 300B tokens from an industrial pre-training corpus containing mixed Chinese, English, mathematics, reasoning, books, papers, and code. We evaluate on a fixed validation set sampled from the same underlying distribution and record token-level losses across training checkpoints so that each contextualized token can be analyzed as a function of optimization step. Because real token trajectories are noisy and need not be monotone, we do not enforce $\Delta_i>0$ in the sigmoid fitting and allow both descending and ascending trajectories, including forgetting-like behavior. Training uses Adam with $\beta_1=0.9$ and $\beta_2=0.95$ and a warmup ratio of 10\% without decay. The step-axis study consumed 133 A100 GPU days overall; in addition to the main runs, we performed two extra runs for token-level analysis in Fig.~\ref{fig:token-level-case}.


\subsection{Step-Axis Results}

\emph{Observation 1: The aggregate loss on the step axis exhibits a clear power-law-like regime after the initial transient stage.} As shown in Fig.~\ref{fig:step-axis} (A), once training enters the stable learning phase, the validation loss decays regularly over optimization steps and follows a power-law pattern.

\emph{Observation 2: Token-wise sigmoid fits accurately recover the aggregate loss trajectory.} Fig.~\ref{fig:step-axis} (A) shows that the average loss implied by the fitted token-level sigmoids closely matches the empirical validation loss, indicating that the dominant token-level learning events are already well captured by the parameterization in Section~\ref{sec:method}. Additional raw token-loss trajectories across learning-time range are provided in Appendix~\ref{app:real-token-trajectories}.

\emph{Observation 3: Token-level learning pulses are highly similar, whereas the learning-time spectrum is strongly heavy-tailed.} Fig.~\ref{fig:step-axis} (B, C) shows that the main cross-token variation lies primarily in \emph{when} tokens are learned, rather than in the detailed local shape of \emph{how} they are learned.

\emph{Observation 4: The empirical loss derivative is governed primarily by the learning-time spectrum.}  In Fig.~\ref{fig:step-axis} (D), the empirical loss derivative, the sigmoid-implied derivative, and the spectrum-dominated reconstruction closely agree, showing that replacing token-specific pulses by a shared pulse leaves the macroscopic derivative shape largely unchanged.

Taken together, these observations establish the full mechanism on the step axis: sigmoid-parameterized token transitions recover the empirical loss dynamics, learning pulses are approximately shared, and the remaining macroscopic shape is governed primarily by the learning-time spectrum. We next test whether the same mechanism continues to hold on the $D$ and $M$ axes.
\begin{figure}[t]
    \centering
    \includegraphics[width=\linewidth]{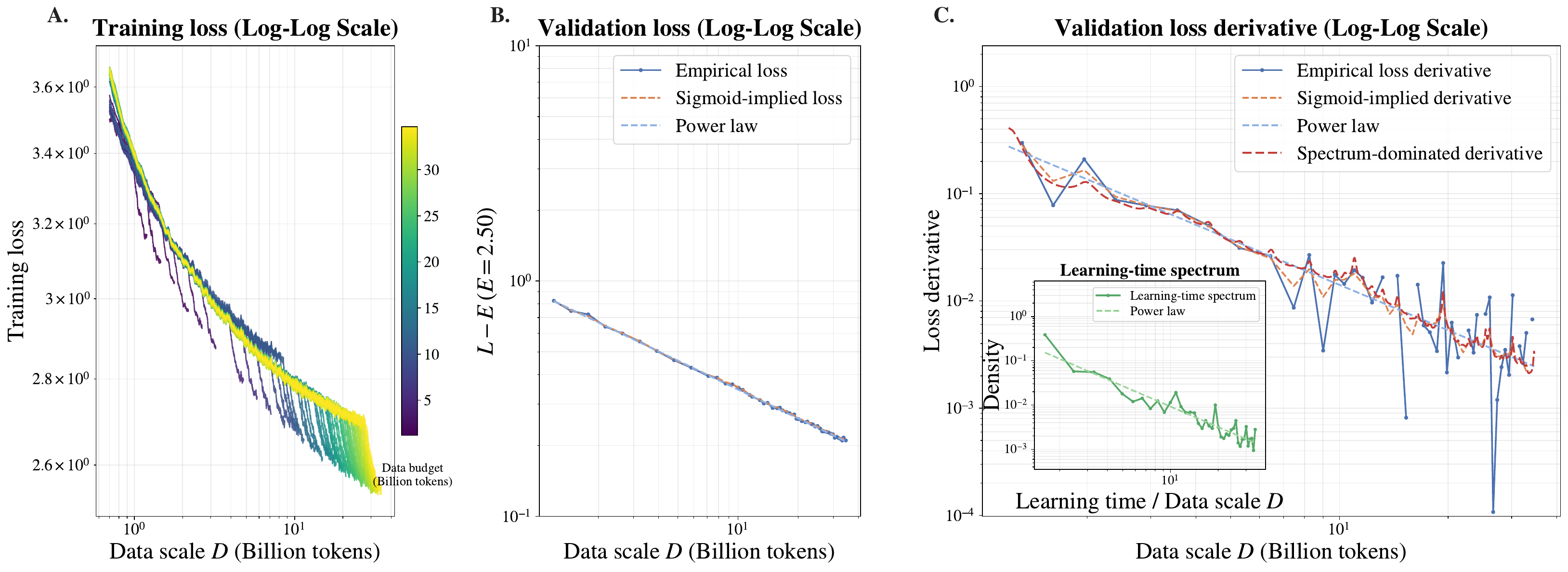}
    \caption{\textbf{A.} Training-loss trajectories from a 50-run data-scale sweep with budgets ranging from 1B to 35B tokens, used to construct the data-scale frontier. \textbf{B.} Frontier validation loss on the $D$ axis together with its sigmoid fit; the resulting envelope remains well described by a power law. \textbf{C.} Learning-time spectrum, empirical loss derivative, sigmoid-implied derivative, power-law fit, and spectrum-dominated reconstruction on the data axis. 
    }
    \label{fig:D-axis}
\end{figure}

\section{Data and Model Axes Decomposition}
\label{sec:nd}

Having established the full decomposition on the step axis, we next ask whether the same pulse-spectrum mechanism also governs the classical scaling-law axes of data scale and model scale. These experiments follow standard scaling-law practice: for each setting, we tune the training configuration to obtain near-optimal performance rather than inheriting a single fixed recipe from the step-axis analysis. The data source remains the same as before, while the scaling variable changes from training-token budget on the $D$ axis to model size on the $M$ axis.

\subsection{Data-Axis Results}

For the data-scale analysis, we fix the model to a 290M MoE architecture and vary the training-token budget from 1B to 35B tokens. Each run uses a complete warmup-constant-decay schedule. To obtain a reliable scaling curve rather than an artifact of suboptimal training choices, we run a large sweep of 50 pre-training experiments and search over training configurations for each budget following standard scaling-law practice. The resulting family of trajectories allows us to construct the envelope validation-loss curve on the $D$ axis, which exhibits a clear power-law trend and serves as the macroscopic object to be explained. This final configuration for the data axis training consumed 358 A100 GPU days in total.

We then perform the same token-level decomposition with data scale as the horizontal axis, fitting each validation token as a function of $D$ and estimating the corresponding learning-time spectrum, shared pulse, and loss-derivative reconstruction. As shown in Fig.~\ref{fig:D-axis}, the same mechanism reappears on the data axis: the token-level sigmoid fits capture the dominant transitions, the learning-time spectrum remains strongly heavy-tailed, and the spectrum-dominated reconstruction closely matches the empirical loss derivative. Thus, even under a frontier constructed from recipe-optimized runs, the macroscopic scaling curve is governed primarily by where learning events are distributed along the data-scale axis, rather than by token-specific variation in local transition shape.

\subsection{Model-Axis Results}

\begin{figure}[t]
    \centering
    \includegraphics[width=\linewidth]{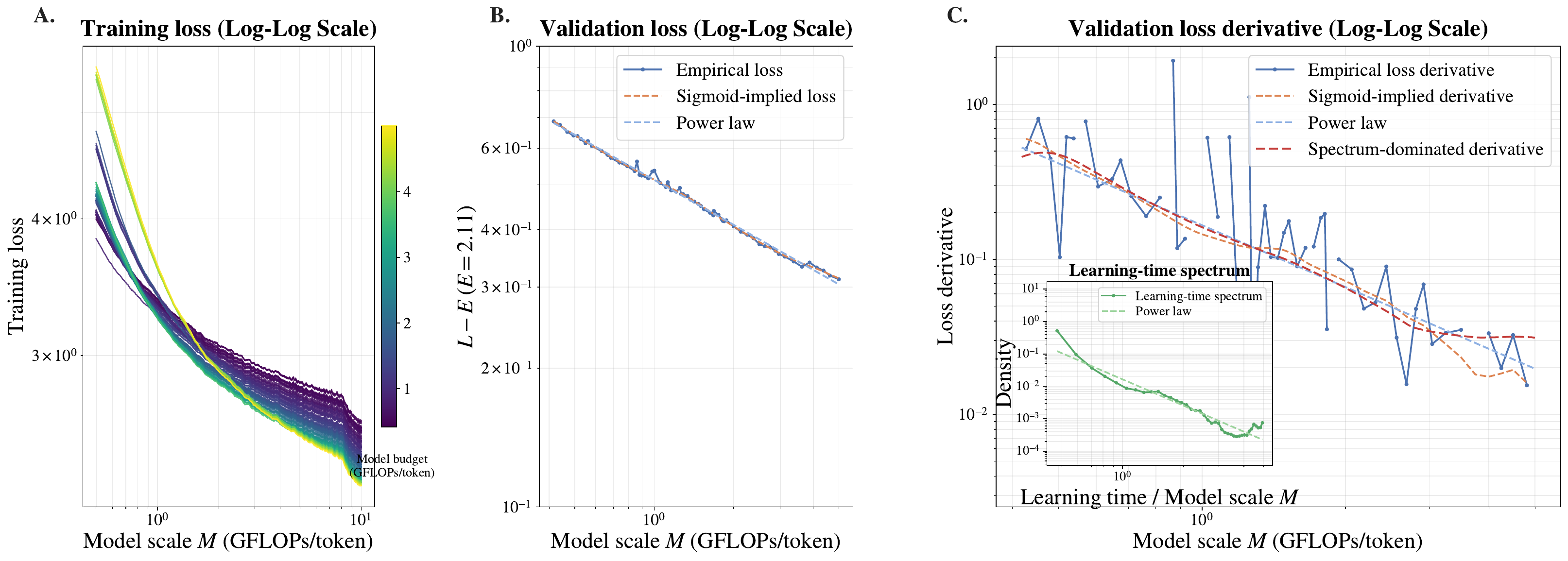}
    \caption{\textbf{A.} Training-loss trajectories from a 60-run model-scale sweep spanning 0.42 to 5.00 GFLOPs/token, used to construct the model-scale frontier. \textbf{B.} Frontier validation loss on the $M$ axis together with its sigmoid fit; the resulting envelope again follows a clear power law. \textbf{C.} Learning-time spectrum, empirical loss derivative, sigmoid-implied derivative, power-law fit, and spectrum-dominated reconstruction on the model-scale axis. 
    }
    \label{fig:M-axis}
\end{figure}

We use non-embedding FLOPs per token $M$ rather than parameter count $N$ to represent model scale. For MoE models, parameter count alone is not a sufficiently faithful measure of effective scale, because it does not reflect the amount of computation and capacity actually activated during training. In particular, as noted in prior work~\cite{bi2024deepseek}, $N$ can introduce substantial distortions when comparing small and medium-scale models. Using $M$ therefore provides a more stable and comparable horizontal axis for scaling-law analysis.

For the model-scale analysis, we fix the training-token budget to 10B tokens and train 60 model variants spanning 0.42 to 5.00 GFLOPs/token, corresponding roughly to 400M to 6B parameters. Following standard scaling-law practice, we tune the architecture and training configuration for each setting to obtain a near-optimal frontier on the $M$ axis. The final configuration of model-axis training consumed 422 A100 GPU days in total. We then perform the same token-level decomposition with model scale as the horizontal axis. As shown in Fig.~\ref{fig:M-axis}, the frontier loss again follows a power law, the learning-time spectrum remains strongly heavy-tailed, and the spectrum-dominated reconstruction closely matches the empirical loss derivative. Thus, the same pulse-spectrum mechanism persists when the horizontal axis is changed from training progress or data scale to model scale.


\section{Reshaping Scaling Behavior}
\label{sec:aux}

\subsection{Static Reshaping with Measured Learning Times}
\begin{figure}[t]
    \centering
    \includegraphics[width=\linewidth]{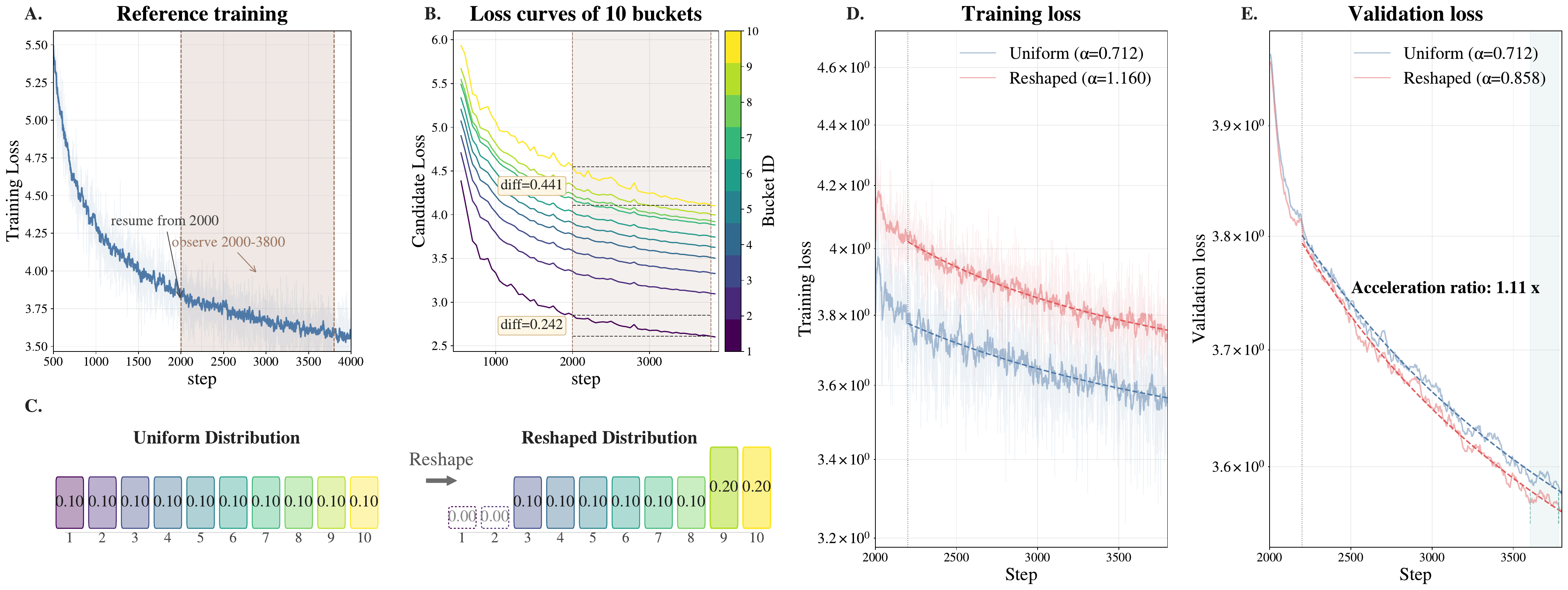}
    \caption{Intervening on the training distribution using the measured learning-time signal. \textbf{A.} We first train a reference model and save checkpoints to estimate learning times for a randomly sampled set of 1B candidate tokens. \textbf{B.} Training samples are then ranked by the fraction of token-level learning events that occur within a target step interval (here, steps 2000--3800) and partitioned into 10 buckets. \textbf{C.} We reshape the training distribution by downweighting the lowest-ranked buckets and upweighting the highest-ranked buckets. \textbf{D.} Continuing training from step 2000 with the reshaped distribution yields a faster decline in training loss than the original distribution, despite a higher irreducible-loss floor. \textbf{E.} The reshaped distribution also improves validation-loss dynamics, achieving a 11\% acceleration relative to the original distribution, which shows that the measured learning-time signal is not only descriptive but can also be used to controllably reshape the scaling curve.}
    \label{fig:reshape}
\end{figure}

The previous sections establish learning time as an explanatory signal for macroscopic loss dynamics. We now ask whether it is also actionable: \emph{can a measured learning-time signal be used to reshape the training distribution and thereby alter subsequent scaling behavior in a controlled way?}

To test this, we first train a 290M MoE reference model and save a sequence of checkpoints together with the corresponding loss trajectories of a randomly sampled set of 1B candidate tokens. We then fit the token-level sigmoid to estimate the learning time of each token. Because autoregressive next-token prediction requires processing the full prefix, token-level intervention is not a practical training unit. We therefore operate at the sample level, which is the smallest unit at which the data distribution can be reshaped efficiently in pre-training. For each sample, we compute a score based on the fraction of its tokens whose estimated learning times fall within a target interval of training steps (here, steps 2000--3800), and rank samples accordingly. The ranked samples are partitioned into 10 buckets, after which we reshape the training distribution by downweighting the lowest-ranked buckets and upweighting the highest-ranked buckets. We then continue training from the beginning of the target interval using the reshaped distribution and compare the resulting training and validation trajectories against the original distribution.

As shown in Fig.~\ref{fig:reshape}, the resulting buckets exhibit clearly ordered loss behavior: higher-ranked samples contain more tokens whose learning events fall within the target interval and therefore undergo larger loss reductions during that phase. Reshaping the training distribution accordingly leads to a faster decline in training loss than the original distribution, indicating that the measured learning-time signal captures an actionable notion of sample learnability. This intervention also improves validation dynamics, yielding a 11\% acceleration relative to the original distribution. These results show that the learning-time signal is not only descriptive but can also be used to intervene on macroscopic scaling behavior in a controlled way.


\subsection{Dynamic Reshaping with Predicted Learnability}
\label{sec:dynamic-reshaping}

\begin{figure}[t]
    \centering
    \includegraphics[width=\linewidth]{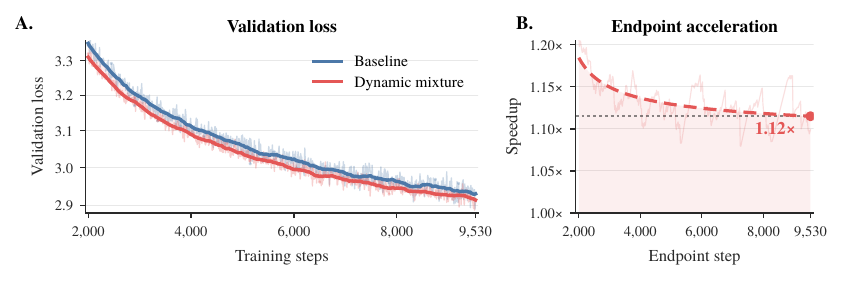}
    \caption{Predictive data reshaping from sample-level loss trajectories. \textbf{A.} Validation loss of the baseline and the dynamic policy, which reallocates sampling probability according to predicted sample learning potential after a 1,000-step burn-in period. \textbf{B.} Equal-loss acceleration measured from policy activation. Predictive reshaping provides sustained gains throughout training and converges to an approximately $1.12\times$ speedup.}
\label{fig:predictive-reshaping}
\end{figure}

The static intervention above establishes that learning-time signals can causally reshape subsequent loss dynamics. We next ask whether they can support a predictive policy that adapts the distribution as the model's learning state evolves. To this end, we model each candidate sample's loss trajectory with a power law and extrapolate its future curve, using the predicted loss reduction as a reusable estimate of its learning potential at each training stage.

We then train a 290M MoE model from scratch for 5B tokens. After a 1,000-step burn-in period, when the initial optimization transient has subsided, we activate a dynamic sampling policy implemented in Megatron. At each subsequent stage, it reallocates probability mass toward samples with the greatest predicted learning potential. All other training configurations are identical to the baseline, isolating the effect of dynamic reshaping.

As shown in Figure~\ref{fig:predictive-reshaping}, predictive reshaping consistently improves validation-loss reduction throughout training. Equal-loss endpoint matching from the activation step yields sustained acceleration and converges to an approximately $1.12\times$ speedup. The gain is therefore not transient: the policy continues to match the distribution to the model's evolving learning state. Moreover, fitting the sample-level curves is a one-time offline cost that can be amortized across repeated runs over the same candidate data pool. We provide a more detailed analysis of stage-dependent distribution reshaping in Appendix~\ref{app:synthetic-reshaping}.

\section{Conclusion}

We introduced a token-level framework for decomposing scaling laws through the distribution of learning times. Across the training-step, data-scale, and model-scale axes, we found that aligned learning pulses are approximately shared and that the macroscopic loss derivative is governed primarily by the learning-time spectrum. This provides direct empirical evidence that the power-law form of language-model loss is determined mainly by \emph{when} tokens are learned. We further showed that the same signal is actionable: reshaping the training distribution with learning-time estimates alters optimization dynamics and improves validation-loss reduction. These results connect token-level learning dynamics to macroscopic scaling behavior and open a path toward data-centric control of large-scale model training.

\section*{Funding and Competing Interests}
\paragraph{Funding.}
This work was supported by Dots Studio, Xiaohongshu Inc. and Shanghai Jiao Tong University.
\paragraph{Competing Interests.}
The authors declare no competing interests.

\bibliographystyle{unsrtnat}
\bibliography{ref}

@article{kaplan2020scaling,
  title={Scaling laws for neural language models},
  author={Kaplan, Jared and McCandlish, Sam and Henighan, Tom and Brown, Tom B and Chess, Benjamin and Child, Rewon and Gray, Scott and Radford, Alec and Wu, Jeffrey and Amodei, Dario},
  journal={arXiv preprint arXiv:2001.08361},
  year={2020}
}

@inproceedings{hoffmann2022training,
  title={Training compute-optimal large language models},
  author={Hoffmann, Jordan and Borgeaud, Sebastian and Mensch, Arthur and Buchatskaya, Elena and Cai, Trevor and Rutherford, Eliza and de Las Casas, Diego and Hendricks, Lisa Anne and Welbl, Johannes and Clark, Aidan and others},
  booktitle={Proceedings of the 36th International Conference on Neural Information Processing Systems},
  pages={30016--30030},
  year={2022}
}

@article{porian2024resolving,
  title={Resolving discrepancies in compute-optimal scaling of language models},
  author={Porian, Tomer and Wortsman, Mitchell and Jitsev, Jenia and Schmidt, Ludwig and Carmon, Yair},
  journal={Advances in Neural Information Processing Systems},
  volume={37},
  pages={100535--100570},
  year={2024}
}

@inproceedings{chen2025revisiting,
  title={Revisiting scaling laws for language models: The role of data quality and training strategies},
  author={Chen, Zhengyu and Wang, Siqi and Xiao, Teng and Wang, Yudong and Chen, Shiqi and Cai, Xunliang and He, Junxian and Wang, Jingang},
  booktitle={Proceedings of the 63rd Annual Meeting of the Association for Computational Linguistics (Volume 1: Long Papers)},
  pages={23881--23899},
  year={2025}
}

@inproceedings{gadre2025language,
  title={Language models scale reliably with over-training and on downstream tasks},
  author={Gadre, Samir Yitzhak and Smyrnis, Georgios and Shankar, Vaishaal and Gururangan, Suchin and Wortsman, Mitchell and Shao, Rulin and Mercat, Jean and Fang, Alex and Li, Jeffrey and Keh, Sedrick and others},
  booktitle={The Thirteenth International Conference on Learning Representations},
  year={2025}
}

@article{hestness2017deep,
  title={Deep learning scaling is predictable, empirically},
  author={Hestness, Joel and Narang, Sharan and Ardalani, Newsha and Diamos, Gregory and Jun, Heewoo and Kianinejad, Hassan and Patwary, Md Mostofa Ali and Yang, Yang and Zhou, Yanqi},
  journal={arXiv preprint arXiv:1712.00409},
  year={2017}
}

@inproceedings{luo2025multi,
  title={A Multi-Power Law for Loss Curve Prediction Across Learning Rate Schedules},
  author={Luo, Kairong and Wen, Haodong and Hu, Shengding and Sun, Zhenbo and Liu, Zhiyuan and Sun, Maosong and Lyu, Kaifeng and Chen, Wenguang},
  booktitle={The Thirteenth International Conference on Learning Representations},
  year={2025}
}

@inproceedings{choshen2025hitchhiker,
  title={A Hitchhiker's Guide to Scaling Law Estimation},
  author={Choshen, Leshem and Zhang, Yang and Andreas, Jacob},
  booktitle={International Conference on Machine Learning},
  pages={10683--10699},
  year={2025},
  organization={PMLR}
}

@inproceedings{bordelon2024dynamical,
  title={A dynamical model of neural scaling laws},
  author={Bordelon, Blake and Atanasov, Alexander and Pehlevan, Cengiz},
  booktitle={Proceedings of the 41st International Conference on Machine Learning},
  pages={4345--4382},
  year={2024}
}

@article{li2025functional,
  title={Functional scaling laws in kernel regression: Loss dynamics and learning rate schedules},
  author={Li, Binghui and Chen, Fengling and Huang, Zixun and Wang, Lean and Wu, Lei},
  journal={arXiv preprint arXiv:2509.19189},
  year={2025}
}

@article{michaud2023quantization,
  title   = {The Quantization Model of Neural Scaling},
  author  = {Michaud, Eric and Liu, Ziming and Girit, Uzay and Tegmark, Max},
  journal = {Advances in Neural Information Processing Systems},
  volume  = {36},
  pages   = {28699--28722},
  year    = {2023}
}

@article{brill2024neural,
  title   = {Neural Scaling Laws Rooted in the Data Distribution},
  author  = {Brill, Ari},
  journal = {arXiv preprint arXiv:2412.07942},
  year    = {2024}
}

@inproceedings{cagnetta2025learning,
  title        = {Learning Curves Theory for Hierarchically Compositional Data with Power-Law Distributed Features},
  author       = {Cagnetta, Francesco and Kang, Hyunmo and Wyart, Matthieu},
  booktitle    = {International Conference on Machine Learning},
  pages        = {6149--6164},
  year         = {2025},
  organization = {PMLR}
}

@article{hutter2021learning,
  title   = {Learning Curve Theory},
  author  = {Hutter, Marcus},
  journal = {arXiv preprint arXiv:2102.04074},
  year    = {2021}
}

@article{maloney2022solvable,
  title   = {A Solvable Model of Neural Scaling Laws},
  author  = {Maloney, Alexander and Roberts, Daniel A. and Sully, James},
  journal = {arXiv preprint arXiv:2210.16859},
  year    = {2022}
}

@article{allen2023physics,
  title={Physics of language models: Part 1, learning hierarchical language structures},
  author={Allen-Zhu, Zeyuan and Li, Yuanzhi},
  journal={arXiv preprint arXiv:2305.13673},
  year={2023}
}

@article{barkeshli2026origin,
  title={On the origin of neural scaling laws: from random graphs to natural language},
  author={Barkeshli, Maissam and Alfarano, Alberto and Gromov, Andrey},
  journal={arXiv preprint arXiv:2601.10684},
  year={2026}
}

@article{chang2024characterizing,
  title={Characterizing Learning Curves During Language Model Pre-Training: Learning, Forgetting, and Stability},
  author={Chang, Tyler A and Tu, Zhuowen and Bergen, Benjamin K},
  journal={Transactions of the Association for Computational Linguistics},
  volume={12},
  pages={1346--1362},
  year={2024}
}

@inproceedings{xia2023training,
  title={Training trajectories of language models across scales},
  author={Xia, Mengzhou and Artetxe, Mikel and Zhou, Chunting and Lin, Xi Victoria and Pasunuru, Ramakanth and Chen, Danqi and Zettlemoyer, Luke and Stoyanov, Veselin},
  booktitle={Proceedings of the 61st Annual Meeting of the Association for Computational Linguistics (Volume 1: Long Papers)},
  pages={13711--13738},
  year={2023}
}

@article{lin2024not,
  title={Not all tokens are what you need for pretraining},
  author={Lin, Zhenghao and Gou, Zhibin and Gong, Yeyun and Liu, Xiao and Shen, Yelong and Xu, Ruochen and Lin, Chen and Yang, Yujiu and Jiao, Jian and Duan, Nan and others},
  journal={Advances in Neural Information Processing Systems},
  volume={37},
  pages={29029--29063},
  year={2024}
}

@inproceedings{carlini2022quantifying,
  title={Quantifying memorization across neural language models},
  author={Carlini, Nicholas and Ippolito, Daphne and Jagielski, Matthew and Lee, Katherine and Tramer, Florian and Zhang, Chiyuan},
  booktitle={The Eleventh International Conference on Learning Representations},
  year={2022}
}

@inproceedings{biderman2023pythia,
  title={Pythia: A suite for analyzing large language models across training and scaling},
  author={Biderman, Stella and Schoelkopf, Hailey and Anthony, Quentin Gregory and Bradley, Herbie and O’Brien, Kyle and Hallahan, Eric and Khan, Mohammad Aflah and Purohit, Shivanshu and Prashanth, USVSN Sai and Raff, Edward and others},
  booktitle={International conference on machine learning},
  pages={2397--2430},
  year={2023},
  organization={PMLR}
}

@inproceedings{michaelov2025language,
  title={Language Model Behavioral Phases are Consistent Across Architecture, Training Data, and Scale},
  author={Michaelov, James A and Levy, Roger P and Bergen, Ben},
  booktitle={The Thirty-ninth Annual Conference on Neural Information Processing Systems},
  year={2025}
}

@article{chen2023skill,
  title={Skill-it! a data-driven skills framework for understanding and training language models},
  author={Chen, Mayee and Roberts, Nicholas and Bhatia, Kush and Wang, Jue and Zhang, Ce and Sala, Frederic and R{\'e}, Christopher},
  journal={Advances in Neural Information Processing Systems},
  volume={36},
  pages={36000--36040},
  year={2023}
}

@inproceedings{nair2024curriculum,
  title={Curriculum learning for small code language models},
  author={Na{\"\i}r, Marwa and Yamani, Kamel and Lhadj, Lynda and Baghdadi, Riyadh},
  booktitle={Proceedings of the 62nd Annual Meeting of the Association for Computational Linguistics (Volume 4: Student Research Workshop)},
  pages={390--401},
  year={2024}
}

@inproceedings{platanios2019competence,
  title={Competence-based curriculum learning for neural machine translation},
  author={Platanios, Emmanouil Antonios and Stretcu, Otilia and Neubig, Graham and Poczos, Barnabas and Mitchell, Tom},
  booktitle={Proceedings of the 2019 conference of the North American chapter of the association for computational linguistics: human language technologies, volume 1 (long and short papers)},
  pages={1162--1172},
  year={2019}
}

@inproceedings{oh2024frequency,
  title={Frequency explains the inverse correlation of large language models’ size, training data amount, and surprisal’s fit to reading times},
  author={Oh, Byung-Doh and Yue, Shisen and Schuler, William},
  booktitle={Proceedings of the 18th Conference of the European Chapter of the Association for Computational Linguistics (Volume 1: Long Papers)},
  pages={2644--2663},
  year={2024}
}

@inproceedings{varre2025learning,
  title={Learning In-context $ n $-grams with Transformers: Sub-$ n $-grams Are Near-Stationary Points},
  author={Varre, Aditya and Y{\"u}ce, Gizem and Flammarion, Nicolas},
  booktitle={International Conference on Machine Learning},
  pages={60924--60963},
  year={2025},
  organization={PMLR}
}

@article{cagnetta2026deriving,
  title={Deriving Neural Scaling Laws from the statistics of natural language},
  author={Cagnetta, Francesco and Ravent{\'o}s, Allan and Ganguli, Surya and Wyart, Matthieu},
  journal={arXiv preprint arXiv:2602.07488},
  year={2026}
}

@article{bi2024deepseek,
  title={Deepseek llm: Scaling open-source language models with longtermism},
  author={Bi, Xiao and Chen, Deli and Chen, Guanting and Chen, Shanhuang and Dai, Damai and Deng, Chengqi and Ding, Honghui and Dong, Kai and Du, Qiushi and Fu, Zhe and others},
  journal={arXiv preprint arXiv:2401.02954},
  year={2024}
}

@article{allen2025physics,
  title={Physics of Language Models: Part 4.1, Architecture design and the magic of Canon layers},
  author={Allen-Zhu, Zeyuan},
  journal={arXiv preprint arXiv:2512.17351},
  year={2025}
}

@inproceedings{rahaman2019spectral,
  title={On the spectral bias of neural networks},
  author={Rahaman, Nasim and Baratin, Aristide and Arpit, Devansh and Draxler, Felix and Lin, Min and Hamprecht, Fred and Bengio, Yoshua and Courville, Aaron},
  booktitle={International conference on machine learning},
  pages={5301--5310},
  year={2019},
  organization={PMLR}
}

@inproceedings{xu2020curriculum,
  title={Curriculum learning for natural language understanding},
  author={Xu, Benfeng and Zhang, Licheng and Mao, Zhendong and Wang, Quan and Xie, Hongtao and Zhang, Yongdong},
  booktitle={Proceedings of the 58th annual meeting of the association for computational linguistics},
  pages={6095--6104},
  year={2020}
}

@article{nam2024exactly,
  title={An exactly solvable model for emergence and scaling laws in the multitask sparse parity problem},
  author={Nam, Yoonsoo and Fonseca, Nayara and Lee, Seok H and Mingard, Chris and Louis, Ard A},
  journal={Advances in Neural Information Processing Systems},
  volume={37},
  pages={39632--39693},
  year={2024}
}

@article{xie2023doremi,
  title={Doremi: Optimizing data mixtures speeds up language model pretraining},
  author={Xie, Sang Michael and Pham, Hieu and Dong, Xuanyi and Du, Nan and Liu, Hanxiao and Lu, Yifeng and Liang, Percy S and Le, Quoc V and Ma, Tengyu and Yu, Adams Wei},
  journal={Advances in Neural Information Processing Systems},
  volume={36},
  pages={69798--69818},
  year={2023}
}

@article{albalak2023efficient,
  title={Efficient online data mixing for language model pre-training},
  author={Albalak, Alon and Pan, Liangming and Raffel, Colin and Wang, William Yang},
  journal={arXiv preprint arXiv:2312.02406},
  year={2023}
}

@article{wei2022emergent,
  title={Emergent abilities of large language models},
  author={Wei, Jason and Tay, Yi and Bommasani, Rishi and Raffel, Colin and Zoph, Barret and Borgeaud, Sebastian and Yogatama, Dani and Bosma, Maarten and Zhou, Denny and Metzler, Donald and others},
  journal={arXiv preprint arXiv:2206.07682},
  year={2022}
}

@article{piantadosi2014zipf,
  title={Zipf’s word frequency law in natural language: A critical review and future directions},
  author={Piantadosi, Steven T},
  journal={Psychonomic bulletin \& review},
  volume={21},
  number={5},
  pages={1112--1130},
  year={2014},
  publisher={Springer}
}

@inproceedings{zhang2026beyond,
  title={Beyond random sampling: Efficient language model pretraining via curriculum learning},
  author={Zhang, Yang and Mohamed, Amr and Abdine, Hadi and Shang, Guokan and Vazirgiannis, Michalis},
  booktitle={Proceedings of the 19th Conference of the European Chapter of the Association for Computational Linguistics (Volume 1: Long Papers)},
  pages={5776--5794},
  year={2026}
}

@inproceedings{zhang2021you,
  title={When do you need billions of words of pretraining data?},
  author={Zhang, Yian and Warstadt, Alex and Li, Xiaocheng and Bowman, Samuel},
  booktitle={Proceedings of the 59th annual meeting of the association for computational linguistics and the 11th international joint conference on natural language processing (volume 1: long papers)},
  pages={1112--1125},
  year={2021}
}

@article{linders2023zipf,
  title={Zipf’s law revisited: Spoken dialog, linguistic units, parameters, and the principle of least effort},
  author={Linders, Guido M and Louwerse, Max M},
  journal={Psychonomic Bulletin \& Review},
  volume={30},
  number={1},
  pages={77--101},
  year={2023},
  publisher={Springer}
}

@inproceedings{wang2026mining,
  title={Mining Useful General Data for Low-Resource Domain Adaptation},
  author={Wang, Pingjie and Liu, Hongcheng and Liao, Yusheng and Fan, Ziqing and Du, Yaxin and Tang, Shuo and Wang, Yanfeng and Wang, Yu},
  booktitle={Proceedings of the 43rd International Conference on Machine Learning},
  year={2026},
  eprint={2511.07380},
  archivePrefix={arXiv},
  primaryClass={cs.CL}
}

@misc{anonymous2026small,
  title={Small Models Know Which Data Scales},
  author={{Anonymous}},
  year={2026},
  note={Under review at the Fifteenth International Conference on Learning Representations},
  url={https://openreview.net/forum?id=gEWI6zkyM8}
}

\newpage
\appendix

\startcontents[appendix]
\section{Appendix}
\printcontents[appendix]{}{1}{\section*{Contents}\setcounter{tocdepth}{3}}

\newpage

\subsection{Why the token distribution in natural language is power-law}
\label{app:token_difficulty_powerlaw}

We now give a compact structural account of why token ``difficulty'' in natural language can be approximately power-law distributed.
The ``difficulty'' of a token is not an intrinsic label of its surface form, but a dynamical quantity: for a token $x$, let $\tau(x)$ be the learning time at which its loss undergoes its main drop.
A token is hard when the regularities needed to predict it are learned late.
Since natural language contains long-tailed linguistic units and usage patterns~\cite{piantadosi2014zipf,linders2023zipf}, the relevant object is not the distribution of tokens alone, but the distribution of learnable regularities that tokens depend on.
Let $\mathcal R$ be the set of latent regularities, and let $\ell(r)\in\{1,2,\dots\}$ denote the compositional level of regularity $r$.
Because natural language is hierarchical and compositional, the number of candidate regularities grows rapidly with level, while each concrete high-level regularity becomes more specific and is reused less often. A minimal abstraction is
\[
N_\ell \propto a^\ell,\qquad
f_\ell \propto q^{-\ell},\qquad a,q>1,
\]
where $N_\ell$ is the number of regularities at level $\ell$ and $f_\ell$ is their typical frequency. Eliminating $\ell$ gives
\[
N(f)\propto f^{-\alpha},\qquad \alpha=\frac{\log a}{\log q},
\]
so a power-law tail arises from the combination of compositional growth in possible patterns and decreasing reuse frequency of each concrete pattern.

To connect this linguistic structure to training dynamics, assume that the time needed to learn a regularity increases with its level or rarity. Let $\tau(r)$ be the learning time of regularity $r$. With an increasing map $T$ and, in the simplest case, exponential growth in learning time,
\[
\tau(r)=T(\ell(r)),\qquad
\tau_\ell \propto s^\ell,\qquad s>1.
\]
Since the cumulative number of regularities up to level $\ell$ scales as $a^\ell$ and $\ell$ grows logarithmically with $\tau$, the induced distribution over regularity learning times is heavy-tailed. Finally, a token becomes easy only after the regularities needed to predict it have been acquired. If $\mathcal R(x)$ denotes those regularities, then
\[
\tau(x)\approx \max_{r\in \mathcal R(x)} \tau(r),
\]
so a heavy-tailed learning-time distribution over regularities transfers to a heavy-tailed, approximately power-law distribution over token learning times. Under this view, token difficulty is the visible projection of hierarchical linguistic structure through the dynamics of model learning.

\subsection{Alignment Between Learning Time and Macroscopic Data Consensus}
\label{app:align-tau-consensus}

\begin{figure}[t]
    \centering
    \includegraphics[width=\linewidth]{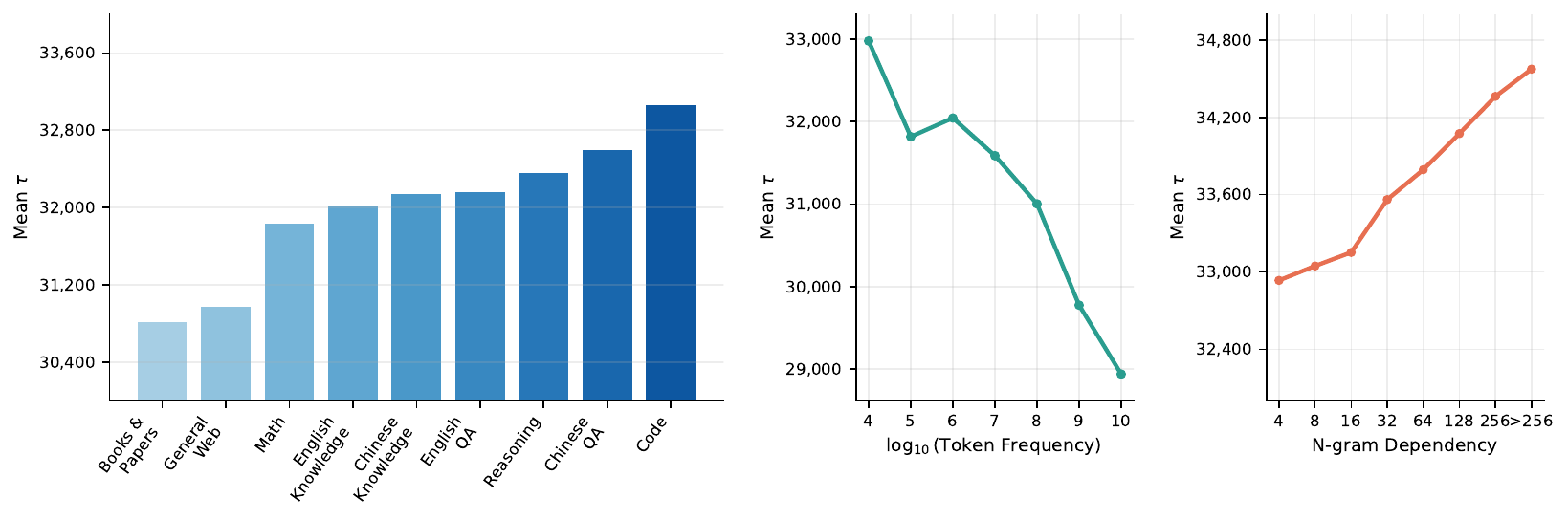}
    \caption{Alignment between token learning time and three macroscopic data regularities.
    \textbf{Left}: mean learning time $\tau$ across validation subsets from different domains. General-text domains such as books, papers, and general web data are learned earlier, while more specialized domains such as math, reasoning, and code are learned later.
    \textbf{Middle}: mean learning time as a function of token frequency bins. Higher-frequency tokens have smaller mean $\tau$, whereas lower-frequency tokens are learned later.
    \textbf{Right}: mean learning time as a function of the context length needed to recover $90\%$ of full-context predictive performance. Token instances that require longer contexts have systematically larger mean $\tau$.}
    \label{fig:apx-align-consensus}
\end{figure}

Many prior studies and industrial practices have formed a set of stable macroscopic intuitions about how language data should be organized during pre-training~\cite{chen2023skill,xie2023doremi,chang2024characterizing}.
Our goal here is to test whether these macroscopic consensus views reappear at the microscopic level as an ordering over token learning times $\tau$.
If our interpretation is correct, then data associated with simpler and more reusable regularities should have smaller mean $\tau$, whereas data associated with rarer, more abstract, or more long-range regularities should have larger mean $\tau$.
Fig.~\ref{fig:apx-align-consensus} shows that this is indeed what we observe.
Even under mixed training, the model does not learn all token groups simultaneously, but instead self-organizes into a learning-time order that closely matches existing intuitions.

\subsubsection{From General Text to Reasoning-Heavy Data}

At the macroscopic level, curriculum-learning studies have repeatedly suggested that reasoning-heavy data should not be injected too early or too aggressively, and that pre-training often benefits from a progression from general text to more specialized domains, as well as from basic syntax to more abstract reasoning~\cite{chen2023skill, nair2024curriculum}.
The left panel of Fig.~\ref{fig:apx-align-consensus} provides a microscopic validation of this consensus through the learning-time spectrum.
When we compute the mean $\tau$ for validation tokens grouped by domain, tokens from books, papers, and general web text appear earliest, while math and knowledge-oriented data lie in the middle, and explicit reasoning, QA, and code tokens appear later.
In particular, code has the largest mean $\tau$, and reasoning-oriented data are also clearly shifted to the right relative to general-language domains.

This ordering suggests that even when all domains are mixed together during training, the model still follows an implicit internal curriculum rather than learning every domain uniformly from the start.
General-language tokens, which mainly reflect common lexical and syntactic structure, tend to be acquired earlier.
By contrast, tokens tied to abstract logic, symbolic manipulation, or domain-specific conventions tend to require later learning events.
We also observe that Chinese tokens have slightly larger mean $\tau$ than English tokens, which may reflect differences in tokenization granularity, corpus balance, and the higher sparsity of some Chinese lexical or domain-specific patterns in the mixed pre-training distribution.

\subsubsection{Token Frequency}

At the macroscopic level, natural language is well known to have a long-tailed frequency structure, and models are widely believed to acquire high-frequency words and shallow patterns before rare tokens and more complex usages~\cite{platanios2019competence, oh2024frequency, chang2024characterizing}.
The middle panel of Fig.~\ref{fig:apx-align-consensus} shows that the learning-time analysis is highly consistent with this view.
For each token, we compute its corpus frequency and then examine how mean $\tau$ changes across frequency bins.
The overall trend is strongly decreasing: high-frequency tokens have smaller mean $\tau$, while low-frequency tokens are learned substantially later.

Although there is mild local fluctuation, the global negative correlation is clear across the entire frequency range.
This means that frequent and widely reused tokens tend to trigger earlier learning events, whereas rare and long-tail tokens tend to populate the late part of the learning-time spectrum.
In this sense, the long tail of token frequency is mirrored by a long tail in $p(\tau)$.
The microscopic distribution of learning times therefore offers a direct dynamical explanation for why rare linguistic material contributes disproportionately to the late-training regime.

\subsubsection{N-gram Regularities and Dependency Length}

At the macroscopic level, previous work has argued that large language models first absorb low-order and short-range $n$-gram regularities, and only later transition to higher-order patterns and longer-range dependencies~\cite{chang2024characterizing, varre2025learning, cagnetta2026deriving}.
To test this microscopically, we assign each token instance a dependency length defined as the minimum context length required to recover $90\%$ of its full-context predictive performance.
The right panel of Figure~\ref{fig:apx-align-consensus} shows a clear monotone trend: token instances that can be recovered from short contexts have smaller mean $\tau$, whereas token instances that require longer contexts have progressively larger mean $\tau$.
The increase continues steadily from short-range cases to the $>256$-token regime.

This pattern indicates that later learning events are not arbitrary, but are systematically associated with token instances whose prediction depends on more extended contextual structure.
In other words, short local regularities tend to be learned earlier, while long-range and higher-order dependencies are learned later.
The familiar macroscopic story of ``short $n$-grams first, long dependencies later'' is therefore visible again at the microscopic level as a rightward shift of mean $\tau$.
Together with the domain and frequency results above, this supports the broader claim that many apparently distinct curriculum effects can be unified as differences in when tokens enter the learned set.

\subsection{Additional Real-Data Validation}
\subsubsection{Token-Level Trajectories}
\label{app:real-token-trajectories}

\begin{figure}[t]
    \centering
    \includegraphics[width=\linewidth]{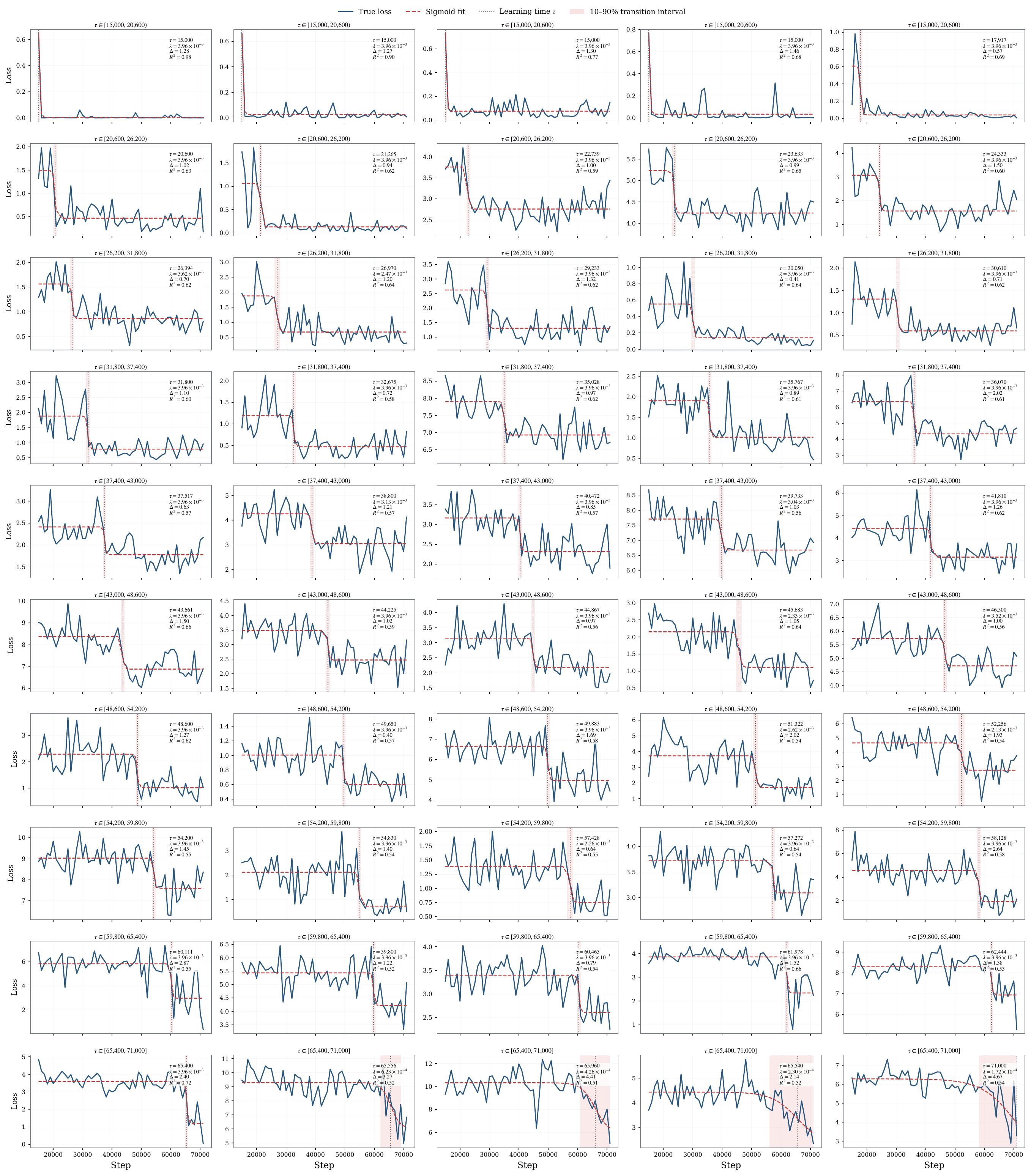}
    \caption{RawEach row represents an equal-width learning-time interval and
    shows five deterministically selected token instances. Blue curves are raw
    checkpoint losses, red dashed curves are fitted sigmoids, gray dotted lines
    mark the fitted learning times $\tau$, and shaded regions indicate the
    10--90\% transition intervals. The examples reveal localized learning events
    throughout real-language pre-training despite substantial token-level noise.}
    \label{fig:apx-real-token-trajectories}
\end{figure}

To complement the representative cross-run cases in Fig.~\ref{fig:token-level-case}, we examine a substantially larger collection of raw contextualized-token loss trajectories from the step-axis experiment. We divide the observed training range into ten learning-time intervals and deterministically select five high-signal examples from each interval, thereby covering learning events throughout the training process. As shown in Fig.~\ref{fig:apx-real-token-trajectories}, despite the substantial noise of real-language pre-training, token losses consistently exhibit a localized transition from a higher-loss regime to a lower-loss regime. These transitions occur at widely different training stages, while the fitted sigmoids accurately identify their locations and dominant loss reductions. Together, these examples provide direct token-level evidence that the smooth aggregate loss curve is composed of asynchronous, localized learning events. For complete inspection and reproducibility, we release the full loss matrix containing 1M token trajectories on the Hugging Face repository\footnote{\url{https://huggingface.co/datasets/applewpj/token-learning-spectrum-examples}}, together with the fitting and analysis code\footnote{\url{https://github.com/applewpj/token-learning-spectrum}}.

\subsubsection{Shared Pulses Across Data Domains}

The domain ordering above shows that different types of data are learned at systematically different times. This raises a complementary question: \emph{do domains also differ in how individual learning events unfold?} Resolving this question is important because the decomposition in Section~\ref{sec:method} uses a shared pulse template $g_{\mathrm{global}}$; averaging across all tokens could otherwise conceal domain-specific pulse shapes. We therefore partition validation tokens into seven data categories and compare three complementary statistics. The median $\Delta_i$ measures the magnitude of a learning event, the median $\lambda_i$ measures its sharpness, and the cosine similarity between the category-average centered pulse $g_c$ and $g_{\mathrm{global}}$ measures its normalized shape. If different domains induce distinct local learning mechanisms, these quantities---especially the pulse similarity---should vary substantially. Conversely, if domains primarily affect when tokens are learned, the pulse shapes should remain stable despite modest parameter variation.

\begin{table}[t]
    \centering
    \caption{Learning-event statistics across data domains. Parentheses report ratios to the corresponding global median. The final row gives the cosine similarity between each category-average centered pulse $g_c$ and the global template $g_{\mathrm{global}}$.}
    \label{tab:apx-domain-pulse-similarity}
    \scriptsize
    \setlength{\tabcolsep}{2.5pt}
    \renewcommand{\arraystretch}{1.15}
    \begin{tabular}{lccccccc}
        \toprule
        & \shortstack{General\\Web} & Math & \shortstack{English\\Knowledge} & \shortstack{Chinese\\Knowledge} & \shortstack{English\\QA} & Reasoning & Code \\
        \midrule
        \shortstack[l]{Median $\Delta_i$\\(global: 0.815)}
        & \shortstack{0.815\\($0.999\times$)}
        & \shortstack{0.803\\($0.985\times$)}
        & \shortstack{0.792\\($0.972\times$)}
        & \shortstack{0.829\\($1.018\times$)}
        & \shortstack{0.796\\($0.976\times$)}
        & \shortstack{0.806\\($0.988\times$)}
        & \shortstack{0.855\\($1.049\times$)} \\
        \shortstack[l]{Median $\lambda_i$\\(global: 0.313)}
        & \shortstack{0.323\\($1.032\times$)}
        & \shortstack{0.324\\($1.035\times$)}
        & \shortstack{0.264\\($0.844\times$)}
        & \shortstack{0.313\\($1.000\times$)}
        & \shortstack{0.298\\($0.952\times$)}
        & \shortstack{0.289\\($0.922\times$)}
        & \shortstack{0.358\\($1.144\times$)} \\
        $\cos(g_c,g_{\mathrm{global}})$
        & 1.000 & 0.999 & 1.000 & 1.000 & 1.000 & 1.000 & 1.000 \\
        \bottomrule
    \end{tabular}
\end{table}

Table~\ref{tab:apx-domain-pulse-similarity} supports the latter prediction. Across domains, the median loss-drop magnitude remains within $0.972$--$1.049\times$ of the global value, while the median sharpness remains within $0.844$--$1.144\times$. Code tokens exhibit somewhat larger and sharper learning events, whereas English-knowledge tokens exhibit smaller and broader events. Despite these differences in scale and width, every category-average pulse has cosine similarity of at least $0.999$ with $g_{\mathrm{global}}$. Thus, data domains differ primarily in \emph{when} their tokens are learned, as captured by their domain-specific learning-time spectra, rather than in the normalized shape of \emph{how} individual learning events unfold. The residual variation in $\Delta_i$ and $\lambda_i$ is secondary and does not invalidate the shared-pulse approximation used in our reconstruction.

\subsubsection{Robustness Across Architectures, Data Mixtures, and Tokenizers}

The main real-data experiments use MoE models with a fixed data mixture and tokenizer, raising the possibility that the observed decomposition is specific to expert routing, a particular corpus composition, or a particular tokenization scheme. To test this alternative, we repeat the real-data analysis with a 19M dense Transformer trained for 3B tokens. We evaluate four combinations of data mixture and tokenizer while keeping the dense-model setup fixed, and measure the $R^2$ between the empirical validation-loss derivative and its pulse-spectrum reconstruction. If the mechanism were configuration-specific, reconstruction quality should deteriorate after removing MoE routing or changing either component of the data pipeline. Conversely, consistently high $R^2$ would indicate that the decomposition captures a more general property of token learning.

\begin{table}[h]
    \centering
    \caption{Dense-model reconstruction across data mixtures and tokenizers. All settings use a 19M dense Transformer trained for 3B tokens. We report the $R^2$ between the empirical validation-loss derivative and its pulse-spectrum reconstruction.}
    \label{tab:apx-dense-robustness}
    \small
    \setlength{\tabcolsep}{12pt}
    \begin{tabular}{lcc}
        \toprule
        Data mixture & Tokenizer & $R^2 \uparrow$ \\
        \midrule
        Original & Qwen~2 & 0.955 \\
        Remixed & Qwen~2 & 0.971 \\
        StarCoder & Qwen~2 & 0.988 \\
        StarCoder & DeepSeek~V3 & 0.955 \\
        \bottomrule
    \end{tabular}
\end{table}

As shown in Table~\ref{tab:apx-dense-robustness}, reconstruction remains strong in every dense-model setting, with $R^2$ ranging from $0.955$ to $0.988$. With the Qwen~2 tokenizer fixed, changing the mixture from the original composition to Remixed or StarCoder yields $R^2$ values of $0.971$ and $0.988$, respectively. With the StarCoder mixture fixed, replacing Qwen~2 with the DeepSeek~V3 tokenizer still gives an $R^2$ of $0.955$. Together with the MoE results in the main text and the dense synthetic experiments in Appendix~\ref{app:synthetic-results}, these findings show that the pulse-spectrum decomposition is not an artifact of MoE routing, the original data mixture, or a particular tokenizer. Rather, within the evaluated settings, it remains accurate across substantially different model and data configurations.

\subsection{Controlled Synthetic Validation}
\label{app:synthetic-results}

\subsubsection{Why Controllable Synthetic Data}

We introduce controllable synthetic data to test the core causal claim of this paper in a setting where the underlying difficulty distribution is explicitly designed rather than only indirectly observed. On natural language corpora, one can measure token-level learning times after training, but it is difficult to intervene on the latent distribution that generates them while holding the rest of the training pipeline fixed. Synthetic data removes this ambiguity: the task family, the difficulty variable, the train-validation split, and the sampling distribution can all be specified a priori, so changes in the global loss curve can be traced back to changes in the learning-time spectrum much more cleanly.

To this end, we build our synthetic benchmark on top of the PhysicsLM synthetic pretraining playground~\cite{allen2023physics, allen2025physics}. We choose PhysicsLM because it provides a deterministic and fully programmable family of algorithmic tasks whose compositional structure is explicit at data-generation time. Among its tasks, we use \emph{Mano}\cite{allen2025physics}, a modular arithmetic reasoning task in which each example is a compositional expression built from addition, subtraction, and multiplication over a finite value space with modulus $23$. The target is the final answer token, whose vocabulary is restricted to tokens $5000$--$5022$. This setup is especially suitable for our purpose because the supervision is concentrated on a single answer token, so each training example induces a sharply defined learning event instead of mixing many heterogeneous supervised positions within one sequence.

The key controllable variable in \emph{Mano} is the expression length, denoted by $L$, which measures the compositional depth of the arithmetic expression and therefore acts as an explicit difficulty variable. Larger $L$ requires more intermediate operations to be composed before the final answer can be predicted, so it naturally induces later and harder learning events. PhysicsLM allows us to specify the sampling distribution over $L$ directly in the data-generation config. This is the central notion of ``control'' in our synthetic setup: we keep the task semantics, tokenizer, model, optimizer, and training procedure fixed, and intervene only on the distribution over compositional difficulty. As a result, any systematic change in the aggregate loss curve can be attributed to a change in the distribution of learning times induced by the data distribution, rather than to a change in architecture or optimization.

\subsubsection{Data Construction and Training Setup}

Following this principle, we construct two training datasets, each containing $51.2$M non-overlapping Mano examples generated in incremental mode from deterministic seeds. The first is a baseline dataset with a uniform difficulty distribution over $L \in \{1,\dots,10\}$. The second is a power-law-skewed dataset whose sampling weights over $L$ are chosen to approximate a discrete distribution proportional to $L^{-2}$ on the same support. We use separate held-out validation sets generated from non-overlapping seed ranges, so training and validation examples are disjoint by construction. In the main synthetic runs, evaluation is performed on a $10$K uniform validation set for diagnosing how learning progresses across difficulty levels.

All synthetic runs use the same training configuration so that the data distribution is the only intended source of variation. We train an $8$-layer, $8$-head transformer with hidden size $512$ and RoPE positional encoding for $400$K optimization steps using Adam with learning rate $3\times 10^{-5}$, batch size $128$, sequence length $64$, BF16 training, and a loss mask that supervises only the final answer token. This last choice is important: because the loss is concentrated on the answer token, the measured learning dynamics reflect when the model becomes able to solve the underlying compositional problem, rather than being diluted by easy prompt tokens.

\subsubsection{Learning Order Under a Uniform Difficulty Distribution}

\begin{figure}[t]
    \centering
    \includegraphics[width=\linewidth]{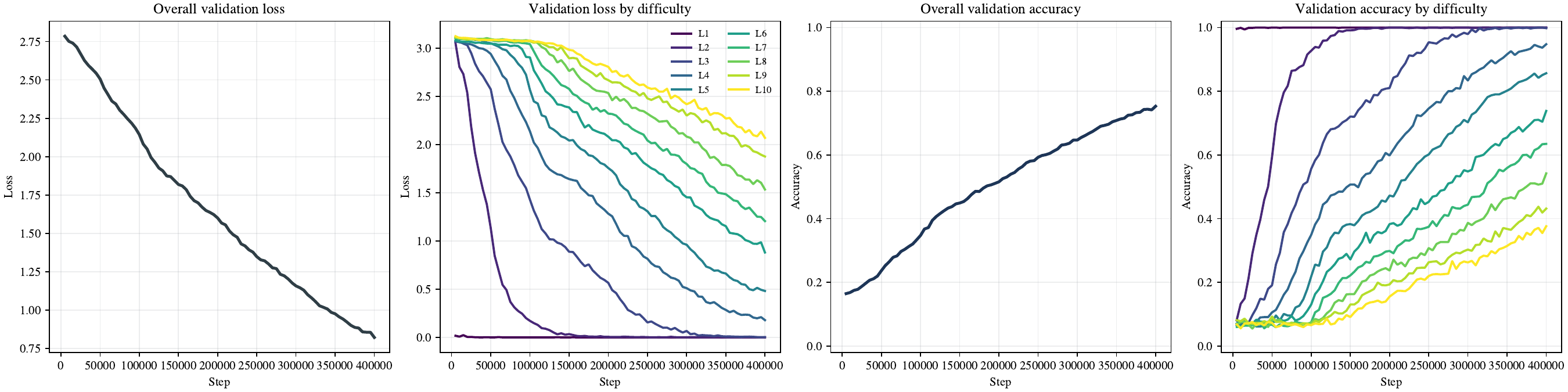}
    \caption{Synthetic learning dynamics under the uniform-\emph{Mano} training distribution. From left to right: overall validation loss, validation loss by difficulty level, overall validation accuracy, and validation accuracy by difficulty level. The aggregate validation loss decreases smoothly but does not show a clear power-law regime; instead, under the uniform difficulty distribution it is much closer to a roughly linear decay over training. At the same time, the per-difficulty curves reveal a strict learning order: low-difficulty samples ($L1$, $L2$, $L3$) are mastered early, whereas higher-difficulty samples improve only later and remain substantially harder throughout training. The same progressive ordering is visible in accuracy as well.}
    \label{fig:apx-synthetic-uniform-learning-order}
\end{figure}

\begin{figure}
    \centering
    \includegraphics[width=\linewidth]{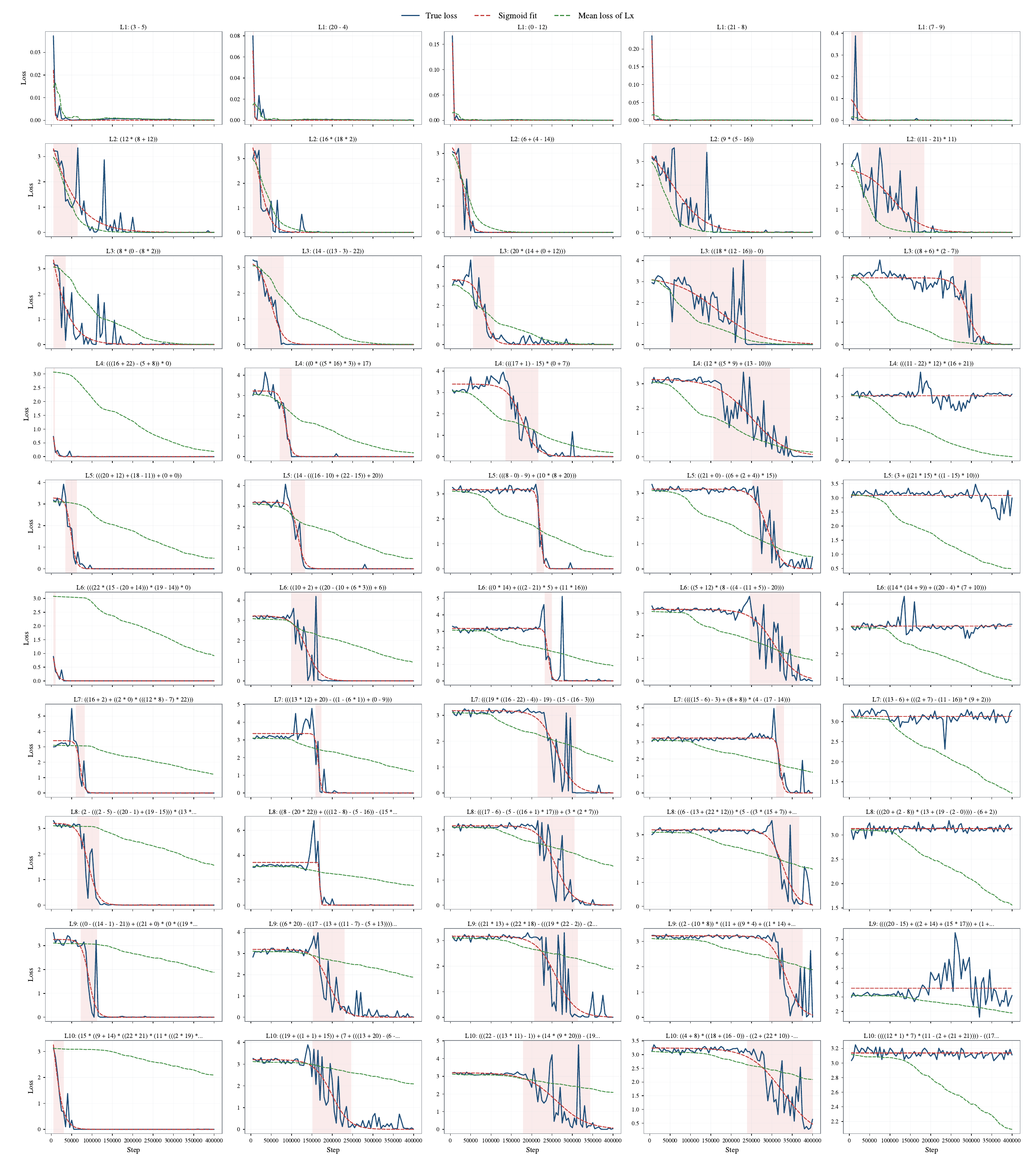}
    \caption{Token-level loss trajectories on uniform-\emph{Mano}. Each row corresponds to one difficulty level from $L1$ to $L10$, and each row shows five sampled tokens from that level. In each panel, the blue curve is the true token loss trajectory over training, the red dashed curve is its sigmoid fit, and the green curve is the average loss trajectory of the corresponding difficulty level. Several regularities are immediately visible. First, except for the easiest $L1$ cases that start decreasing before the main observation window, most tokens begin from a nearly common high-loss state and end near zero, as expected for a deterministic single-answer task. Second, individual token learning is concentrated in a localized transition interval rather than being uniformly distributed over training. Third, different tokens mainly vary in the timing and sharpness of this transition, while the sigmoid fit captures the dominant step-like behavior well. These observations motivate modeling training as a collection of token-level learning events and decomposing aggregate loss dynamics in terms of their distribution over learning times.}
    \label{fig:apx-synthetic-uniform-token-gallery}
\end{figure}

Before turning to token-level fitting, we first verify two basic empirical properties of the uniform-\emph{Mano} setting as shown in Figure~\ref{fig:apx-synthetic-uniform-learning-order}. First, because the training distribution over difficulty levels is uniform rather than power-law, the aggregate validation loss does not exhibit the clear power-law shape observed in real-language pre-training. Instead, it decreases in a much more nearly linear manner over the main training regime. This point is important for our overall argument: the synthetic experiment is not designed to reproduce a power law by construction, but to show that when the underlying difficulty distribution is changed, the shape of the aggregate learning curve changes accordingly.

Second, the learning dynamics are strongly ordered by difficulty. When we evaluate the model separately on validation subsets of different expression lengths, we observe a clear progression from $L1$ to $L10$: the easiest examples are learned first, and the hardest examples are learned last. The same ordering appears consistently in both validation loss and validation accuracy. This shows that the synthetic data indeed contains a meaningful difficulty hierarchy, and that learning over this hierarchy is highly non-uniform rather than simultaneous across all levels. In other words, different parts of the data distribution enter the learned set at different times, which is precisely the condition needed for a nontrivial learning-time spectrum to emerge.

\subsubsection{Token-Level Learning Dynamics}

We next examine token-level loss trajectories under the same uniform-\emph{Mano} setting. As showin in Figure~\ref{fig:apx-synthetic-uniform-token-gallery}, because each sample has a unique correct answer token and the loss is computed only on that answer position, the token-level dynamics are especially clean. Before a token is learned, the model is close to guessing among a fixed and small answer vocabulary, so most tokens start from nearly the same initial loss plateau. Empirically this common starting point is visible for almost all difficulty levels except $L1$, whose easiest tokens often begin to decrease so early that part of the initial drop already occurs before the main observation window. At the other end, once the model has learned the corresponding rule and places essentially all probability mass on the unique correct answer, the terminal loss is naturally driven to zero.

More importantly, the trajectories show that learning is not spread uniformly across the entire training horizon. Instead, each token stays near an initial steady state, then undergoes a relatively sharp transition within a localized interval, and finally enters a second steady state near zero. Different tokens do not differ mainly in whether they exhibit such a transition, but in when this transition happens and how sharp it is. This is exactly the token-level signature of a learning event: a token contributes meaningfully to the aggregate loss decrease only around its own transition window, while contributing little before it is learned or after it has already saturated. From this perspective, the macroscopic loss curve is not a smooth primitive object, but the superposition of many asynchronous token-level learning events occurring at different times.

\subsubsection{Token-level Learning Pulse}

\begin{figure}[t]
    \centering
    \includegraphics[width=\linewidth]{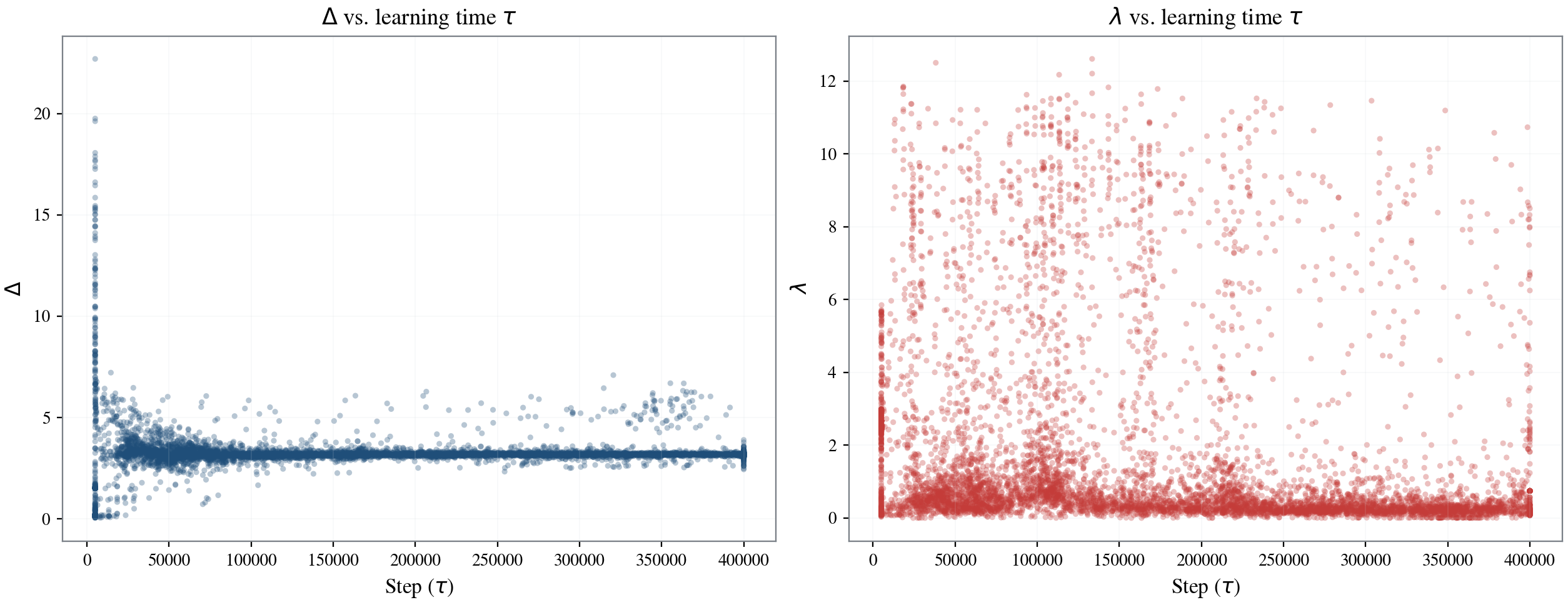}
    \caption{Scatter plots of token-level sigmoid parameters against learning time on uniform-\emph{Mano}. Left: loss-drop magnitude $\Delta_i$ versus learning time $\tau_i$. Right: sharpness parameter $\lambda_i$ versus learning time $\tau_i$. Both quantities vary across tokens, showing that different tokens do not share exactly the same transition amplitude or speed. However, neither $\Delta_i$ nor $\lambda_i$ exhibits a strong systematic dependence on $\tau_i$, indicating that later-learned tokens are not qualitatively different from earlier-learned ones in a way that would by itself determine the global loss shape. This supports the view that the dominant macroscopic factor is the distribution of learning times rather than a strong temporal drift in token-specific pulse shape.}
    \label{fig:apx-synthetic-uniform-tau-params}
\end{figure}

We then fit a sigmoid to every validation token and extract its three token-level parameters $(\tau_i,\lambda_i,\Delta_i)$, corresponding respectively to learning time, transition sharpness, and loss-drop magnitude. These quantities summarize when each token is learned, how quickly the transition happens, and how much loss it removes once learned. Having obtained these parameters for all validation tokens, we next ask whether the macroscopic loss shape is driven mainly by variation in the token-specific pulse shape itself, or by the distribution of learning times across tokens.

We first examine the relationship between the fitted shape parameters and learning time. As shown in Figure~\ref{fig:apx-synthetic-uniform-tau-params}, both $\lambda_i$ and $\Delta_i$ vary across tokens, meaning that different tokens indeed have different drop magnitudes and different transition speeds. However, neither quantity shows a strong systematic trend with $\tau_i$. In other words, although token-level learning events are not literally identical, their local shape variation is not organized primarily along the learning-time axis. This already suggests that the timing distribution may be more important than the detailed per-token pulse parameters in explaining the global loss trajectory.

\begin{figure}[t]
    \centering
    \includegraphics[width=\linewidth]{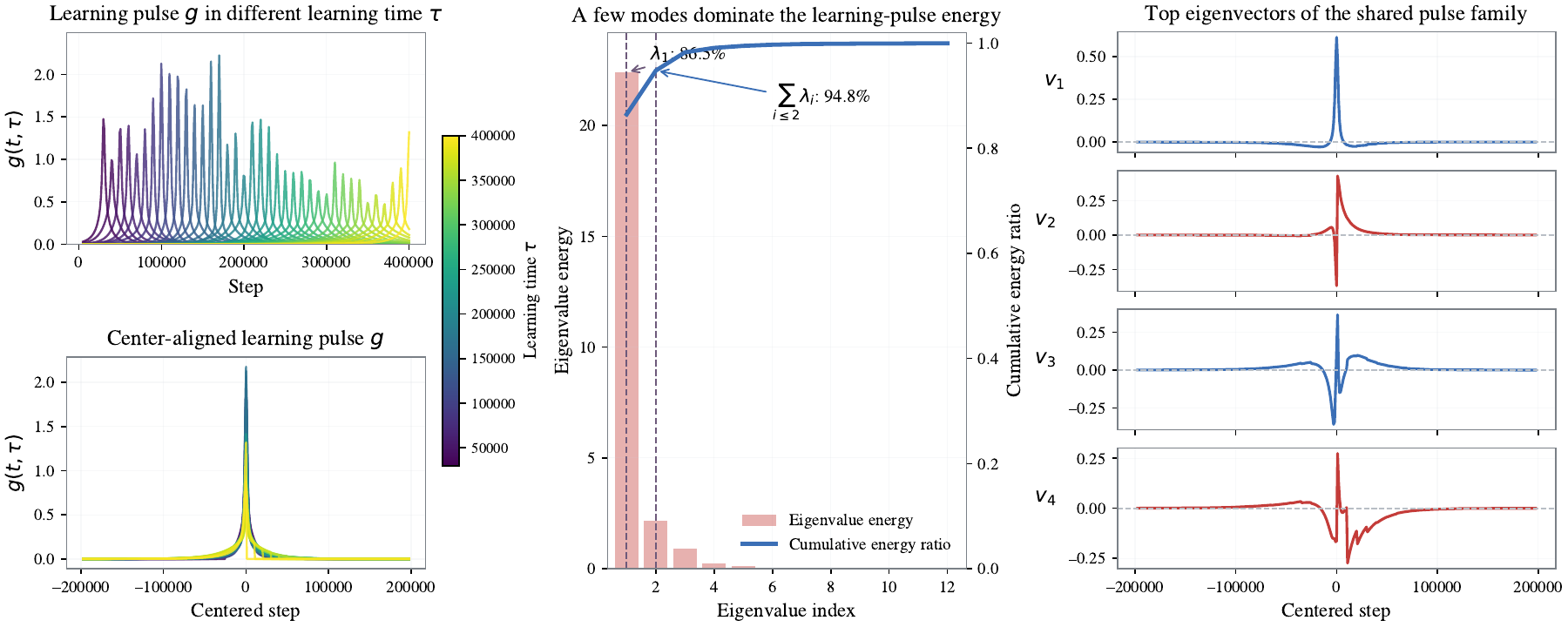}
    \caption{Similarity of learning pulses across learning times on uniform-\emph{Mano}. \textbf{Left}: the upper panel shows the averaged learning pulse for different learning-time groups on the original training axis, while the lower panel shows the same pulses after aligning their centers. The strong collapse after alignment indicates that pulses learned at different times have nearly the same shape up to translation. \textbf{Middle}: singular-value spectrum of the aligned pulse matrix, together with the cumulative energy ratio; the first singular mode explains about $86.3\%$ of the energy and the first two modes explain about $94.8\%$, showing that the aligned pulse family is strongly low-rank. \textbf{Right}: the top four singular vectors of the aligned pulse matrix. The leading mode closely matches the canonical pulse shape, while the higher modes account only for small residual variation. Together these results justify replacing token-specific pulses by a shared average template $g$ when reconstructing aggregate loss dynamics.}
    \label{fig:apx-synthetic-uniform-g-similarity}
\end{figure}

We next visualize the learning pulse induced by tokens at different learning times and align them by their centers. Figure~\ref{fig:apx-synthetic-uniform-g-similarity} shows that before alignment, the pulses appear at different positions along the training axis simply because different token groups are learned at different times. Once centered, however, these pulses almost collapse onto the same shape, revealing a high degree of invariance across learning times. We further quantify this by performing SVD on the matrix of center-aligned learning pulses. The resulting spectrum is strongly low-rank: the first singular mode alone explains about $86.3\%$ of the energy, and the first two modes together explain about $94.8\%$. Moreover, the leading singular vector itself has essentially the same localized shape as the aligned learning pulse family. This provides direct evidence that, after averaging over individual token noise, learning pulses at different times are nearly the same object up to translation.

This observation is central for the decomposition developed in the main text. Although individual tokens can still have visibly different trajectories, their average learning pulse is highly stable once aligned by learning time. We can therefore use a shared template $g$ to represent the typical token-level learning event, and decouple the macroscopic loss decrease from the detailed shape of each token's drop. Under this view, the aggregate loss dynamics are governed primarily by how many tokens undergo this shared learning event at each time, namely by the distribution of learning times along the training axis.

\subsubsection{From Learning-Time Spectrum to Loss Decrease Rate}

Having established that token-level trajectories are well fit by sigmoids and that the aligned learning pulses are highly similar, we can now test the full reconstruction implied by our framework. The key question is whether the learning-time spectrum $p(\tau)$ already captures the dominant structure of the macroscopic loss decrease rate, once the token-specific pulse shape is replaced by a shared average template.

\begin{figure}[t]
    \centering
    \includegraphics[width=\linewidth]{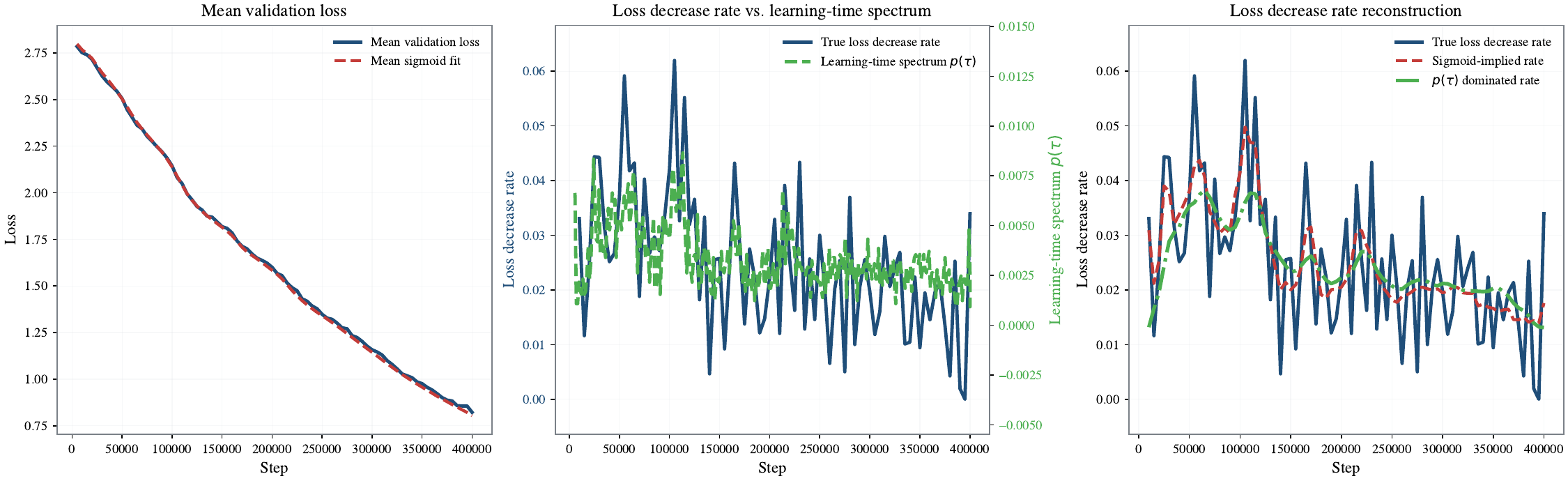}
    \caption{Reconstruction of synthetic aggregate dynamics from token-level sigmoid fits. \textbf{Left}: the true mean validation loss and the mean loss obtained by averaging token-wise sigmoid fits. The two curves are nearly indistinguishable, showing that the sigmoid decomposition faithfully captures the aggregate loss trajectory and that the dominant token-level learning events are already encoded by the fitted parameters. \textbf{Middle}: the true loss decrease rate and the empirical learning-time spectrum $p(\tau)$ on the same step axis. Their shapes are closely matched, indicating that the temporal distribution of learning events already tracks the main structure of macroscopic loss reduction. \textbf{Right}: comparison among the true loss decrease rate, the rate implied by the mean sigmoid fit, and the rate reconstructed from the shared average pulse $g$ together with $p(\tau)$. All three curves nearly overlap, and the reconstruction using the averaged pulse is effectively a smoothed version of the true rate. This shows that $p(\tau)$ captures the principal features of the loss decrease rate, and that once a shared pulse template is fixed, the aggregate rate is determined primarily by how learning events are distributed over time rather than by detailed token-specific pulse variations.}
    \label{fig:apx-synthetic-uniform-reconstruction}
\end{figure}

Figure~\ref{fig:apx-synthetic-uniform-reconstruction} verifies this prediction directly. We first compare the true mean validation loss with the average of all token-wise sigmoid fits. The two curves are almost identical, showing that the fitted sigmoid family already provides an accurate description of the aggregate synthetic loss curve. This means the decomposition is not merely qualitatively plausible, but quantitatively faithful at the level of the overall loss.

We then place the empirical learning-time spectrum $p(\tau)$ and the loss decrease rate on the same step axis. Their distributions are closely matched: the times at which many tokens are learned coincide with the times at which the aggregate loss falls most rapidly. This is exactly the behavior predicted by the pulse-spectrum view, in which the rate of macroscopic improvement is governed by how densely token-level learning events are distributed over time.

Finally, using the shared average learning pulse together with $p(\tau)$, we reconstruct the loss decrease rate and compare it with both the true rate and the rate implied by the full sigmoid decomposition. All three curves nearly overlap, and the reconstruction from the averaged pulse is essentially a smoothed version of the true loss decrease rate. This shows that the dominant structure of aggregate loss reduction is determined mainly by the learning-time spectrum $p(\tau)$. By contrast, the precise shape of $g$ is secondary: once learning pulses are broadly similar, macroscopic loss decay is governed primarily by when tokens learn, not by the fine details of how each individual pulse is shaped.

\subsubsection{Power-Law Synthetic Distribution}

We finally repeat the synthetic analysis under a power-law data distribution to test whether the conclusions above depend on the uniform-difficulty setting. Concretely, we construct both the training and validation sets from the same \emph{Mano} task family, but replace the uniform length distribution by a power-law distribution over difficulty levels. Both the training and validation sets are drawn from the probability mass over $L \in \{1,\dots,10\}$ is chosen to approximate a discrete law proportional to $L^{-2}$. As before, the task definition, model architecture, optimizer, and answer-only supervision remain unchanged, so the only intended intervention is the distribution of compositional difficulty.

\begin{figure}[t]
    \centering
    \includegraphics[width=\linewidth]{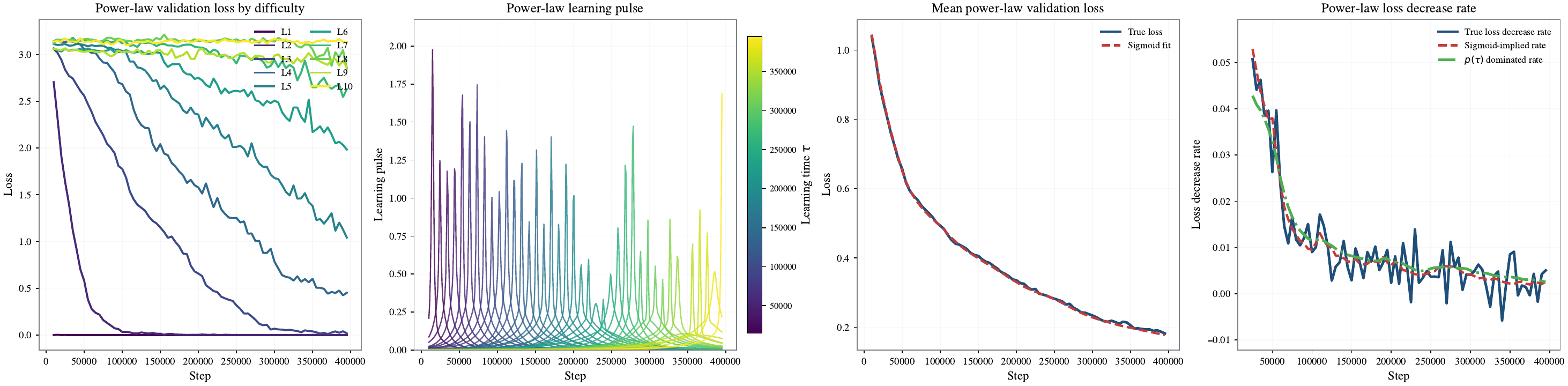}
    \caption{Overview of synthetic learning dynamics under the power-law \emph{Mano} distribution. \textbf{First}: validation loss by difficulty level. Even under the power-law data distribution, the losses of different difficulty levels still exhibit a clear learning order, with easier levels improving earlier and harder levels improving later. \textbf{Second}: learning pulses associated with different learning times. The pulse family remains localized and qualitatively similar across the training axis. \textbf{Third}: mean validation loss under the power-law distribution, showing the true loss together with the sigmoid-implied loss; the two curves remain closely matched. \textbf{Fourth}: validation-loss decrease rate under the power-law distribution, comparing the true decrease rate, the sigmoid-implied decrease rate, and the rate reconstructed from the averaged learning pulse together with $p(\tau)$. The close agreement among these curves shows that the pulse-spectrum decomposition continues to hold in the power-law setting as well.}
    \label{fig:apx-synthetic-powerlaw-overview}
\end{figure}

Figure~\ref{fig:apx-synthetic-powerlaw-overview} shows that the main conclusions from the uniform synthetic setting remain valid after switching to a power-law data distribution. First, the by-difficulty validation losses still display a clear progression from easy to hard levels, so the synthetic task continues to induce an ordered hierarchy of learning events rather than simultaneous learning across all difficulties. Second, the learning pulses associated with different learning times remain sharply localized and visually similar, indicating that the shared-pulse approximation is not an artifact of the uniform-distribution experiment.

The aggregate-level conclusions also persist. The mean validation loss is again accurately captured by the average of token-wise sigmoid fits, showing that the token-level decomposition remains quantitatively faithful. Likewise, the true loss decrease rate, the sigmoid-implied rate, and the rate reconstructed from the averaged learning pulse together with $p(\tau)$ still track one another closely. This means that even when the underlying data distribution itself is power-law, the same mechanism continues to explain the macroscopic loss curve: the dominant structure of loss reduction is determined mainly by the distribution of learning times, while the detailed token-specific pulse shape plays a comparatively secondary role.

\subsection{Learning-Time-Guided Synthetic Data Reshaping}
\label{app:synthetic-reshaping}

Having established that the learning-time spectrum explains aggregate loss dynamics, we ask: \textbf{\emph{How should the training distribution be reshaped, using only a single empirical measurement of learning time, to maximize subsequent learning efficiency?}} We use the reference learning time $\tau_{\mathrm{ref}}$ as a run-conditioned proxy for example difficulty and resample training examples while holding all other training settings fixed.

We organize our analysis around three questions:
\begin{itemize}[leftmargin=*]
    \item \textbf{\emph{Can a learning time measured in a single reference run serve as a reusable difficulty signal?}} This is necessary because any subsequent reshaping strategy assumes that $\tau_{\mathrm{ref}}$ preserves a meaningful example ordering when the training distribution changes.
    
    \item \textbf{\emph{Which examples should be upsampled at a given training stage?}} Resolving this question is essential for obtaining the largest possible gain: indiscriminate reweighting may be ineffective, whereas examples aligned with the model's current learning stage may provide substantially greater utility.
    
    \item \textbf{\emph{Why does effective data reshaping accelerate learning?}} Answering this question distinguishes whether individual token-level learning events become stronger or whether more learning events occur earlier, thereby testing the mechanism predicted by the learning-time-spectrum view.
\end{itemize}

\subsubsection{Learning Order Is Globally Stable}

\begin{figure}[t]
    \centering
    \includegraphics[width=0.5\linewidth]{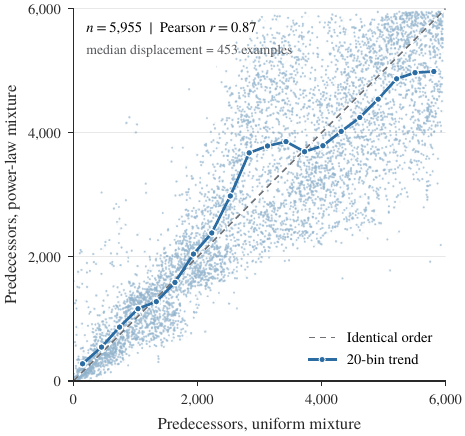}
    \caption{\textbf{Learning order is globally stable but locally variable.}
    For validation examples learned under both the uniform and power-law training mixtures, we compare predecessor counts---the number of shared examples learned earlier in each run. The scatter and 20-bin trend largely follow the identity line (Pearson $r=0.87$), while variability increases for later-learned examples. Thus, a single reference run provides a stable global ordering but uncertain
    local ranks.}
    \label{fig:apx-synthetic-order-stability}
\end{figure}

A prerequisite for learning-time-guided reshaping is that the ordering induced by a single reference run remains informative when the training distribution changes. We train two models for 400,000 steps on the same synthetic task, using either the uniform or the power-law mixture described in Section~\ref{app:synthetic-results}. Throughout training, both models are evaluated on the same held-out validation set of 10,000 examples uniformly distributed over difficulty levels L1--L10, yielding one loss trajectory per run for every validation example. We fit these trajectories to estimate the corresponding learning times and retain the 5,955 examples whose learning events fall within the observed checkpoint range in both runs. Since changing the training mixture can alter the overall pace of learning, we compare relative learning order rather than absolute learning times. For each retained example, we define its \emph{predecessor count} in a run as the number of retained examples learned before it, and compare the two predecessor counts for the same example in the scatter plot in Figure~\ref{fig:apx-synthetic-order-stability}.

If a single-run learning time provides a reusable ordering of example difficulty, the same validation example should occupy a similar relative position under the two training mixtures. We therefore expect its two predecessor counts to be close, producing a scatter concentrated around the diagonal and a binned trend that approximately follows the identity line. By contrast, if learning order is determined primarily by the training distribution, the predecessor counts should show little alignment across runs.

Figure~\ref{fig:apx-synthetic-order-stability} is largely consistent with the first prediction. The individual examples exhibit a clear diagonal structure, the 20-bin trend follows the identity line over most of the range, and the two predecessor counts have a Pearson correlation of $0.87$. At the same time, the scatter around the diagonal becomes progressively wider for later-learned examples, indicating increasing uncertainty in their precise local ordering. These results show that \textbf{learning order is globally stable but locally variable}. The broad progression from earlier- to later-learned examples is preserved across training mixtures, while individual examples exhibit local run-to-run fluctuations that become more pronounced later in training. We therefore treat $\tau_{\mathrm{ref}}$ as a coarse ordering signal rather than an exact sample-level property, motivating the bin-level interventions used below.

\subsubsection{Stage-Dependent Data Utility}
\begin{figure}[t]
    \centering
    \includegraphics[width=\linewidth]{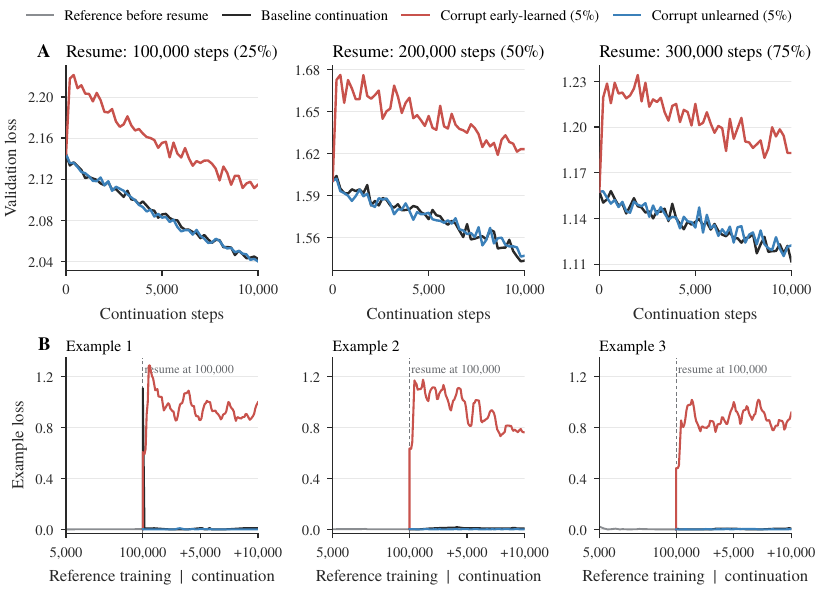}
    \caption{\textbf{Examples outside the current learning region contribute little to subsequent loss reduction.} To identify where probability mass can be reallocated, training is resumed for 10,000 steps from three reference checkpoints using a reference-matched baseline distribution, or variants in which 5\% of the continuation set is replaced by answer-corrupted reference-early-learned or reference-unlearned examples. \textbf{A:} Corrupting reference-unlearned examples produces a trajectory nearly identical to the baseline, whereas corrupting reference-early-learned examples causes an upward loss offset while preserving an approximately parallel subsequent trajectory. \textbf{B:} Representative examples resumed from step 100,000 confirm that the offset arises from forgetting of previously learned behavior. Thus, reference-unlearned examples are effectively indistinguishable from noise at the current stage, while early-learned examples affect the retained loss level but contribute little to the subsequent rate of improvement.}
    \label{fig:apx-synthetic-stage-utility}
\end{figure}

Having established that $\tau_{\mathrm{ref}}$ provides a reusable coarse ordering, we next ask which parts of this ordering are useful at a given training stage. We partition the candidate training pool into a reference-early-learned category with $\tau_{\mathrm{ref}}<5{,}000$, 100 equal-width reference-learning-time bins spanning the observed range, and a reference-unlearned category containing examples with no detected learning event by step 400,000. From a much larger disjoint candidate pool, we construct a baseline continuation set whose composition over these categories matches that of the reference training distribution. We then resume training for 10,000 steps from the 100,000- (25\%), 200,000- (50\%), and 300,000- (75\%) step checkpoints.

\paragraph{Where to Remove Probability Mass.}

Upsampling a selected learning-time bin while keeping the total sampling probability fixed requires removing the same probability mass from another part of the distribution. We therefore first ask which regions provide useful signal at the current training stage and which can be reduced with minimal short-term effect. As a conservative stress test, we replace the answers of examples amounting to 5\% of the continuation set with random incorrect answers, drawing them either from the reference-early-learned or reference-unlearned category. If a category provides little usable signal over the 10,000-step continuation window, its corruption should leave validation dynamics close to the baseline; otherwise, validation performance should deteriorate.

Figure~\ref{fig:apx-synthetic-stage-utility} shows that both extremes can serve as sources of probability mass, but for different reasons. Corrupting reference-unlearned examples leaves the validation trajectory nearly indistinguishable from the baseline, indicating that these examples are effectively no more informative than noise at the current training stage. Corrupting reference-early-learned examples causes previously learned validation examples to be forgotten, producing an upward loss offset; however, the subsequent loss trajectory remains approximately parallel to the baseline. Thus, early-learned examples affect the retained loss level but contribute little to the subsequent rate of improvement, suggesting that cleanly downweighting them need not slow continued learning. In the following experiments, we therefore reallocate probability mass from the reference-early-learned category while preserving the reference-unlearned category, since changing the training distribution may cause some currently unlearned examples to become learnable earlier.

\paragraph{Where to Add Probability Mass.}

\begin{figure}[t]
    \centering
    \includegraphics[width=\linewidth]{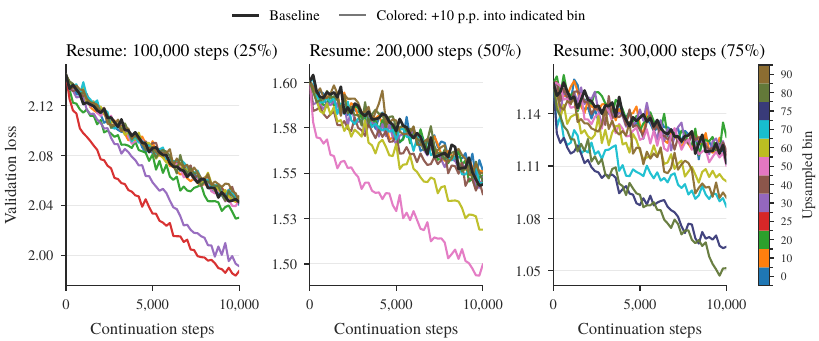}
    \caption{\textbf{Upsampling currently learnable examples accelerates training.} Each colored curve transfers 10 percentage points of sampling probability from the reference-early-learned category into one reference-learning-time bin. If data utility is aligned with the current training stage, the largest improvement should come from bins near the resume checkpoint. Consistent with this prediction, the useful region shifts from bins 25--30 at step 100,000 to bins 50--60 at step 200,000 and bins 70--90 at step 300,000, while distant bins remain close to the baseline. Thus, the most useful data region moves forward with training rather than remaining fixed.}
\label{fig:apx-synthetic-bin-sweep}
\end{figure}

Once a source of probability mass has been identified, the remaining question is where that mass should be reassigned to obtain the largest gain. The value of the reference learning-time signal lies not merely in ranking examples by difficulty, but in identifying which examples are best matched to the model's current learning stage. If this interpretation is correct, upsampling should be most effective for examples whose reference learning times lie near the resume checkpoint, while examples far behind or far ahead should provide little short-term benefit.

Figure~\ref{fig:apx-synthetic-bin-sweep} confirms this stage-local pattern. The most effective region moves from bins 25--30 when training resumes at step 100,000 (25\%), to bins 50--60 at step 200,000 (50\%), and to bins 70--90 at step 300,000 (75\%). Bins far from the corresponding resume checkpoint remain close to, or occasionally worse than, the baseline. The useful region also becomes broader later in training, consistent with the increasing local uncertainty of late learning-time order observed above. These results show that data utility is localized and moves forward along the reference learning-time axis, implying that an effective reshaping strategy must track the model's current training stage rather than apply a fixed difficulty mixture throughout training.

\subsubsection{More Tokens Learning Earlier Drive Acceleration}

\begin{figure}[t]
    \centering
    \includegraphics[width=\linewidth]{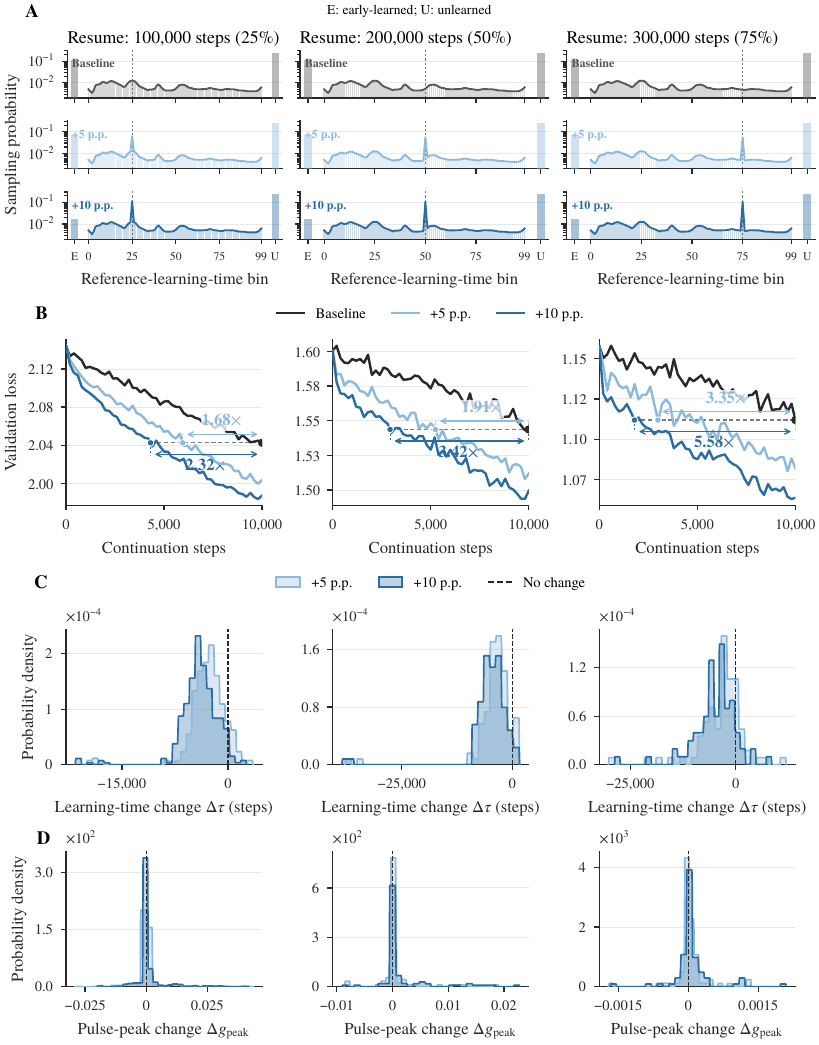}
    \caption{\textbf{Targeted upsampling makes more tokens learn earlier.} \textbf{A:} Sampling distributions for the baseline, +5 p.p., and +10 p.p. interventions. \textbf{B:} The +10 p.p. intervention yields time-to-target speedups of $2.32\times$, $3.42\times$, and $5.58\times$ at the 100,000-, 200,000-, and 300,000-step resumes, revealing larger gains later in training. \textbf{C:} $\Delta\tau$ is predominantly negative for shared validation examples, indicating earlier learning. \textbf{D:} $\Delta g_{\mathrm{peak}}$ remains centered near zero. Thus, the gain comes from earlier rather than stronger learning events.}
\label{fig:apx-synthetic-acceleration-mechanism}
\end{figure}

Having identified which examples should receive additional probability mass, we finally ask why this intervention accelerates learning. Under the learning-time-spectrum view, faster aggregate loss reduction can arise through two distinct mechanisms: individual learning events may become stronger, or learning events may occur earlier so that more tokens have been learned by any fixed training step. Distinguishing these mechanisms is essential for determining whether data reshaping changes the strength of token acquisition or primarily changes when acquisition occurs. We therefore compare each targeted intervention with its matched baseline through both the fitted learning-time shift, $\Delta\tau$, and the change in learning-pulse peak, $\Delta g_{\mathrm{peak}}$. If acceleration is driven by stronger individual events, $\Delta g_{\mathrm{peak}}$ should shift systematically upward. Conversely, if it is driven by more tokens learning earlier, $\Delta\tau$ should shift below zero while $\Delta g_{\mathrm{peak}}$ remains approximately unchanged.

Figure~\ref{fig:apx-synthetic-acceleration-mechanism} supports the latter mechanism. Transferring 10 percentage points of sampling probability into the post-resume target bin yields time-to-target speedups of $2.32\times$, $3.42\times$, and $5.58\times$ when training is resumed at steps 100,000, 200,000, and 300,000, respectively, showing that the gain increases at later training stages. For validation examples learned under both the baseline and intervention, the distributions of $\Delta\tau$ are strongly shifted below zero, with the larger intervention producing a greater shift. In contrast, the corresponding $\Delta g_{\mathrm{peak}}$ distributions remain concentrated around zero without a consistent directional change. Thus, targeted upsampling does not primarily strengthen individual learning events; it advances their occurrence, causing more tokens to cross their learning transitions within the same training budget. Effective data reshaping therefore accelerates learning by shifting the learning-time spectrum toward earlier steps rather than by amplifying the typical learning pulse.

\clearpage

\subsection{Limitations and Future Work}
This work makes a substantial step toward explaining the origin of neural scaling laws by resolving aggregate power-law behavior into token-level learning events and their learning-time spectrum, thereby providing a more mechanistic understanding of why macroscopic loss follows such regular forms. Despite this progress, several important directions remain open for future work. First, the current analysis is computationally expensive, as it requires training-recipe sweeping, dense checkpointing, and token-level trajectory fitting across multiple scaling axes, which currently limits broader coverage over larger experimental grids and additional model families. Second, our current data-reshaping experiments operate at the sample level rather than the token level, since autoregressive next-token prediction couples each target token with its preceding context during the forward pass. The present intervention should therefore be viewed as an initial demonstration that the measured learning-time signal is actionable, while finer-grained token-level reshaping and more complete curriculum-style scheduling remain important directions for future work.

\section*{NeurIPS Paper Checklist}

\begin{enumerate}

\item {\bf Claims}
    \item[] Question: Do the main claims made in the abstract and introduction accurately reflect the paper's contributions and scope?
    \item[] Answer: \answerYes{} 
    \item[] Justification: The abstract and introduction have clearly stated the main claims.

\item {\bf Limitations}
    \item[] Question: Does the paper discuss the limitations of the work performed by the authors?
    \item[] Answer: \answerYes{} 
    \item[] Justification: The paper discusses several limitations.

\item {\bf Theory assumptions and proofs}
    \item[] Question: For each theoretical result, does the paper provide the full set of assumptions and a complete (and correct) proof?
    \item[] Answer: \answerNA{} 
    \item[] Justification: The paper does not include theoretical results.

    \item {\bf Experimental result reproducibility}
    \item[] Question: Does the paper fully disclose all the information needed to reproduce the main experimental results of the paper to the extent that it affects the main claims and/or conclusions of the paper (regardless of whether the code and data are provided or not)?
    \item[] Answer: \answerYes{} 
    \item[] Justification: All the experimental details are provided in Section~\ref{sec:step}, Section~\ref{sec:nd}, Section~\ref{sec:aux}, and the Appendix.

\item {\bf Open access to data and code}
    \item[] Question: Does the paper provide open access to the data and code, with sufficient instructions to faithfully reproduce the main experimental results, as described in supplemental material?
    \item[] Answer: \answerYes{} 
    \item[] Justification: The code and data are open access, and all the core claims of this paper are very straightforward to reproduce with the instructions provided in the paper.

\item {\bf Experimental setting/details}
    \item[] Question: Does the paper specify all the training and test details (e.g., data splits, hyperparameters, how they were chosen, type of optimizer) necessary to understand the results?
    \item[] Answer: \answerYes{} 
    \item[] Justification: All the experimental detailes are provided in Section~\ref{sec:step}, Section~\ref{sec:nd}, Section~\ref{sec:aux}, and the Appendix.

\item {\bf Experiment statistical significance}
    \item[] Question: Does the paper report error bars suitably and correctly defined or other appropriate information about the statistical significance of the experiments?
    \item[] Answer: \answerYes{} 
    \item[] Justification: We repeat the same decomposition process on multiple runs and different axes, which strongly supports the consistency of the mechanism we propose.

\item {\bf Experiments compute resources}
    \item[] Question: For each experiment, does the paper provide sufficient information on the computer resources (type of compute workers, memory, time of execution) needed to reproduce the experiments?
    \item[] Answer: \answerYes{} 
    \item[] Justification: The paper has provided the compute recources for all experiments.
    
\item {\bf Code of ethics}
    \item[] Question: Does the research conducted in the paper conform, in every respect, with the NeurIPS Code of Ethics \url{https://neurips.cc/public/EthicsGuidelines}?
    \item[] Answer: \answerYes{} 
    \item[] Justification: The paper conforms with the NeurIPS Code of Ethics.

\item {\bf Broader impacts}
    \item[] Question: Does the paper discuss both potential positive societal impacts and negative societal impacts of the work performed?
    \item[] Answer: \answerNA{} 
    \item[] Justification: This work helps to understand the mechanism behind scaling laws, which is not intended to have direct societal impacts.
    
\item {\bf Safeguards}
    \item[] Question: Does the paper describe safeguards that have been put in place for responsible release of data or models that have a high risk for misuse (e.g., pre-trained language models, image generators, or scraped datasets)?
    \item[] Answer: \answerNA{} 
    \item[] Justification: The paper poses no such risks.

\item {\bf Licenses for existing assets}
    \item[] Question: Are the creators or original owners of assets (e.g., code, data, models), used in the paper, properly credited and are the license and terms of use explicitly mentioned and properly respected?
    \item[] Answer: \answerYes{} 
    \item[] Justification: All the assets used in the paper are properly credited and the license and terms of use are explicitly mentioned and properly respected.

\item {\bf New assets}
    \item[] Question: Are new assets introduced in the paper well documented and is the documentation provided alongside the assets?
    \item[] Answer: \answerYes{} 
    \item[] Justification: This paper does not introduce new code and data.

\item {\bf Crowdsourcing and research with human subjects}
    \item[] Question: For crowdsourcing experiments and research with human subjects, does the paper include the full text of instructions given to participants and screenshots, if applicable, as well as details about compensation (if any)? 
    \item[] Answer: \answerNA{} 
    \item[] Justification: The paper does not involve crowdsourcing nor research with human subjects.

\item {\bf Institutional review board (IRB) approvals or equivalent for research with human subjects}
    \item[] Question: Does the paper describe potential risks incurred by study participants, whether such risks were disclosed to the subjects, and whether Institutional Review Board (IRB) approvals (or an equivalent approval/review based on the requirements of your country or institution) were obtained?
    \item[] Answer: \answerNA{} 
    \item[] Justification: The paper does not involve crowdsourcing nor research with human subjects.

\item {\bf Declaration of LLM usage}
    \item[] Question: Does the paper describe the usage of LLMs if it is an important, original, or non-standard component of the core methods in this research? Note that if the LLM is used only for writing, editing, or formatting purposes and does \emph{not} impact the core methodology, scientific rigor, or originality of the research, declaration is not required.
    \item[] Answer: \answerNo{} 
    \item[] Justification: The usage of LLMs is not an important, original, or non-standard component of the core methods in this research.

\end{enumerate}

\end{document}